%% file: ms.tex
\documentclass[1p]{elsarticle}

\usepackage{lineno,hyperref}
\modulolinenumbers[5]

\usepackage{graphicx}
\usepackage{subcaption}
\usepackage{trackchanges}
\usepackage{color}
\usepackage{url}
\usepackage{enumitem}
\usepackage{rotating}

\addeditor{Marc}

\newcommand{\acha}{ACHA}

\definecolor{darkgreen}{rgb}{0.05, 0.6, 0.05}

\definecolor{orange}{rgb}{1,0.5,0}

\journal{Artificial Intelligence}

\begin{document}

\begin{frontmatter}

\title{The Hanabi Challenge: A New Frontier for AI Research}
\author[deepmind]{Nolan Bard\corref{equ}}
\ead{nbard@google.com}
\author[oxford]{Jakob N. Foerster\corref{equ}}
\ead{jnf@fb.com}
\author{Sarath Chandar\fnref{brain}}
\ead{chandar@google.com}
\author{Neil Burch\fnref{deepmind}}
\ead{burchn@google.com}
\author{Marc Lanctot\fnref{deepmind}}
\ead{lanctot@google.com}
\author{H. Francis Song\fnref{deepmind}}
\ead{songf@google.com}
\author{Emilio Parisotto\fnref{cmu}}
\ead{eparisot@cs.cmu.edu}
\author{Vincent Dumoulin\fnref{brain}}
\ead{vdumoulin@google.com}
\author{Subhodeep Moitra\fnref{brain}}
\ead{smoitra@google.com}
\author{Edward Hughes\fnref{deepmind}}
\ead{edwardhughes@google.com}
\author{Iain Dunning\fnref{deepmind}}
\ead{iaindunning@gmail.com}
\author{Shibl Mourad\fnref{deepmind}}
\ead{shibl@google.com}
\author{Hugo Larochelle\fnref{brain}}
\ead{hugolarochelle@google.com}
\author{Marc G. Bellemare\fnref{brain}}
\ead{bellemare@google.com}
\author{Michael Bowling\fnref{deepmind}}
\ead{bowlingm@google.com}

\cortext[equ]{Equal Contribution}
\fntext[deepmind]{DeepMind}
\fntext[oxford]{University of Oxford, work done at DeepMind}
\fntext[brain]{Google Brain}
\fntext[cmu]{Carnegie Mellon University, work done at Google Brain}

\begin{abstract}
From the early days of computing, games have been important testbeds for studying
how well machines can do sophisticated decision making. In recent years,
machine learning has made dramatic advances with artificial agents reaching
superhuman performance in challenge domains like Go, Atari, and some
variants of poker.  As with their predecessors of chess, checkers, and backgammon,
these game domains have driven research by providing sophisticated yet
well-defined challenges for artificial intelligence practitioners.  We continue this tradition by proposing the game of Hanabi as a new challenge
domain with novel problems that arise from its combination of purely
cooperative gameplay with two to five players and imperfect information.
In particular, we argue that Hanabi elevates reasoning about the beliefs and
intentions of other agents to the foreground.  We believe developing novel
techniques for such theory of mind reasoning will
not only be crucial for success in Hanabi, but also in broader
collaborative efforts, especially those with human partners.  To facilitate
future research, we introduce the open-source Hanabi Learning Environment,
propose an experimental framework for the research community to evaluate
algorithmic advances, and assess the performance of current state-of-the-art
techniques.
\end{abstract}


\end{frontmatter}


\input{core/1_introduction}

\input{core/2_the_game}
\input{core/5_the_benchmark}
\input{core/6_empirical_evidence}
\input{core/7_related_work}
\input{core/8_conclusion}
\input{core/acks}

\bibliography{hanabi-challenge}
\newpage
\appendix
\section{}

\input{core/appendix}
\end{document}

%% file: core/1_introduction.tex
\section{Introduction}

Throughout human societies, people engage in a wide range of activities with a
diversity of other people. These multi-agent interactions are integral to
everything from mundane daily tasks, like commuting to work, to operating the
organizations that underpin modern life, such as governments and economic
markets.  With such complex multi-agent interactions playing a pivotal role in
human lives, it is desirable for artificially intelligent agents to
also be capable
of cooperating effectively with other agents, particularly humans.

Multi-agent environments present unique challenges relative to those with a
single agent. In particular, the ideal behaviour for an agent typically depends
on how the other agents act. Thus, for an agent to maximize its utility in such
a setting, it must consider how the other agents will behave, and respond
appropriately.  Other agents are often the most complex part of the
environment: their policies are commonly stochastic, dynamically changing, or
dependent on private information that is not observed by everyone.  Furthermore,
agents generally need to interact while only having a limited time to observe
others.

While these issues make inferring the behaviour of others a
daunting challenge for AI practitioners, humans routinely make such inferences
in their social interactions using \emph{theory of
  mind}
~\citep{PremackWoodruff78,Rabinowitz18}: reasoning about others as agents
with their own mental states -- such as perspectives, beliefs, and intentions
-- to explain and predict their behaviour.
\footnote{Dennett~\cite{Dennett87} uses the
  phrase \emph{intentional stance} to refer to this ``strategy'' for
  explanation and prediction.}
Alternatively, one can think of
theory of mind as the human ability to imagine the world from another person's
point of view.  For example, a simple real-world use of theory of mind can be
observed when a pedestrian crosses a busy street. Once some traffic has
stopped, a driver approaching the stopped cars may not be able to directly
observe the pedestrian. However, they can reason about why the other drivers
have stopped, and infer that a pedestrian is crossing.

In this work, we examine the popular card game Hanabi, and argue for it as a
new research frontier that, at its very core, presents the kind of multi-agent
challenges where humans employ theory of mind. Hanabi won the prestigious
\emph{Spiel des Jahres} award in 2013 and enjoys an active community,
including a number of sites that allow for online gameplay~\citep{BoardGameArena,HanabiLive}.  Hanabi is a
\emph{cooperative} game of \emph{imperfect information} for two to five
players, best described as a type of team solitaire.  The game's
imperfect information arises from each player being unable to see their own
cards (\emph{i.e.} the ones they hold and can act on), each of which has a
color and rank.  To succeed, players must coordinate to efficiently reveal
information to their teammates, however players can only 
communicate though \emph{grounded} hint actions that point out all of a player's
cards of a chosen rank or colour.
Importantly, performing a hint action consumes the limited resource
of \emph{information tokens}, making it impossible to fully resolve each
player's uncertainty about the cards they hold based on this grounded
information alone.  For AI practitioners, this restricted communication
structure also prevents the use of ``cheap talk'' communication channels
explored in previous multi-agent research~\citep{Foerster16RIAL,Lewis17,Cao18}.
Successful play involves communicating extra
information implicitly through the choice of actions themselves, which are
observable by all players.

Hanabi is different from the adversarial two-player zero-sum games where
computers have reached super-human skill, \emph{e.g.}, chess~\citep{campbell2002deep}, checkers~\citep{schaeffer1996chinook}, go~\citep{silver16mastering},
backgammon~\citep{tesauro95temporal} and two-player poker~\citep{Moravcik17DeepStack,Brown17Libratus}.  In those games, agents typically compute
an equilibrium policy (or equivalently, a strategy) such that no single player
can improve their utility by deviating from the equilibrium.  While two-player
zero-sum games can have multiple equilibria, different equilibria are
\emph{interchangeable}: each player can play their part of different
equilibrium profiles without impacting their utility. As a result, agents can
achieve a meaningful worst-case performance guarantee in these domains by
finding any equilibrium policy.  However, since Hanabi is neither
(exclusively) two-player nor zero-sum, the value of an agent's policy depends critically on the policies
used by its teammates.  Even if all players manage to play according to the
same equilibrium, there can be multiple locally optimal equilibria that are
relatively inferior.\footnote{One such equilibrium occurs when players do not
  intentionally communicate information to other players, and ignore
  what other players tell them (historically called a pooling equilibrium in
  pure signalling games~\cite{lewis2008convention}, or a babbling
  equilibrium in later work using cheap talk~\cite{Crawford82}).  In this case,
  there is no incentive for a player to start communicating because they will
  be ignored, and there is no incentive to pay attention to other players
  because they are not communicating.}
For algorithms that iteratively train independent agents, such as those
commonly used in the multi-agent reinforcement learning literature, these inferior
equilibria can be particularly difficult to escape
and so even learning a good policy for all players is challenging.

The presence of imperfect information in Hanabi creates another challenging
dimension of complexity for AI algorithms. As has been observed
in domains like poker, imperfect information entangles how an agent should
behave across multiple observed states~\cite{Burch14CFRD,Burch17PhD}. In
Hanabi, we observe this when thinking of the policy as a \emph{communication protocol} \footnote{In pure
signalling games where actions are purely communicative, policies are often
referred to as \emph{communication protocols}.  Though Hanabi is not such a
pure signalling game, when we want to emphasize the communication properties of
an agent's policy we will still refer to its communication protocol.} between players, where
the efficacy of any given protocol depends on the entire scheme rather
than how players communicate in a particular observed situation.  That is, how
the other players will respond to a chosen signal will depend upon what other
situations use the same signal.  Due to this entanglement, the type of
single-action exploration techniques common in reinforcement learning
(\emph{e.g.}, $\epsilon$-greedy, entropy regularization) can incorrectly
evaluate the utility of such exploration steps as they ignore their holistic
impact.

Humans appear to be approaching Hanabi differently than most multiagent
reinforcement learning approaches.  Even beginners with no experience
will start signalling playable cards, reasoning that their teammates'
perspective precludes them from knowing this on their own.
Furthermore, beginners will confidently play cards that are only
partially identified as playable, recognizing that the intent in the
partial identification is sufficient to fully signal its playability.
This all happens on the first game, suggesting players are considering
the perspectives, beliefs, and intentions of the other players (and
expecting the other players are doing the same thing about them).
While hard to quantify, it would seem that theory of mind is a central
feature in how the game is first learned.  We can see further evidence
of theory of mind in the descriptions of advanced
conventions\footnote{%
We use the word \emph{convention} to refer to the parts of a communication
protocol or policy that interrelate.  Technically, these can be thought of as
constraints on the policy to enact the convention.  Conventions can communicate
information either with or without an explicit signal, viz., conventions not only
specify situations when a signal is used and how a teammate responds to that
signal, but the absence of a signal also indicates that players are not in one
of these situations.   While we will also refer
to the conventions of a learned policy, note that this is merely a convenient
abstraction to aid discussion.  The learning agents we examine learn a policy
without explicitly encoding conventions.  More examples of human
conventions in Hanabi will be discussed in Section 2.}
used by
experienced players.  The descriptions themselves often include the
rationale behind each ``agreement'' explicitly including reasoning
about other players' beliefs and intentions.
\begin{quote}
C should assume that D is going to play their yellow card.  C must do
something, and so they ask themselves: ``Why did B give that clue?''.
The only reason is that C can actually make that card playable.~\cite{IraciHanabiConventions}
\end{quote}
Such conventions then enable further reasoning about other players'
beliefs and intentions.  For example, the statement that ``C should
assume that D is going to play their yellow card'', is itself the
result of reasoning that partial identification of a playable card is
sufficient to identify it as playable.

From human play we can also see that the goal itself is multi-faceted.  One
challenge is to learn a policy for the entire team that has high
utility.  Most of the prior AI research on Hanabi has focused on this
challenge, which we refer to as the \emph{self-play} setting.  Human
players will often strive toward this goal, pre-coordinating their
behaviour either explicitly using written guides or implicitly through
many games of experience with the same players.  As one such guide
states, though, ``Hanabi is very complicated, so it is impossible to write a
guide on how to best solve each individual situation.''\cite{ZamiellHanabiConventions}.
Even if if such a guide existed it is impractical for human Hanabi
players to memorize nuanced policies or expect others to do the
same.  However, humans also routinely play with \emph{ad-hoc teams}
that may have players of different skill levels and little or no
pre-coordination amongst everyone on the team.  Even without agreeing
on a complete policy or a set of conventions, humans are still able to achieve a
high degree of success.  It appears that human
efforts in both goals are aided by theory of mind reasoning, and AI
agents with similar capabilities --- playing well in both
pre-coordinated self-play and in uncoordinated ad-hoc teams --- would
signal a useful advance for the field.

The combination of cooperation, imperfect information, and limited
communication make Hanabi an ideal challenge domain for learning in both the
self-play and ad-hoc team settings.
In Section~\ref{sec:hanabi} we describe the details of the game and
how humans approach it.
In Section~\ref{sec:benchmark} we present the Hanabi Learning Environment open source code framework 
(Section~\ref{sec:open-environment}) and guidelines for evaluating both the
self-play (Section~\ref{sec:benchmark:self-play}) and ad-hoc team
(Section~\ref{sec:benchmark:ad-hoc}) settings.   We evaluate the performance of
current state-of-the-art reinforcement learning methods in
Section~\ref{sec:experiments}.  Our results show that although these learning
techniques can achieve reasonable performance in self-play, they generally fall
short of the best known hand-coded agents
(Section~\ref{sec:experiments:self-play}).  Moreover, we show that these
techniques tend to learn extremely brittle policies that are unreliable for ad-hoc
teams (Section~\ref{sec:experiments:ad-hoc}).  These results suggest that there
is still substantial room for technical advancements in both the self-play and
ad-hoc settings, especially as the number of players increases.  Finally, we
highlight connections to prior work in Section~\ref{sec:related}.

%% file: core/2_the_game.tex
\section{Hanabi: The Game}
\label{sec:hanabi}

Hanabi is a game for two to five players, best described as a type of
cooperative solitaire. Each player holds a \emph{hand} of four cards (or five, when
playing with two or three players). Each card depicts a rank (\textbf{1} to
\textbf{5}) and a colour (\textbf{red}, \textbf{green}, \textbf{blue},
\textbf{yellow}, and \textbf{white}); the deck (set of all cards) is composed
of a total of 50 cards, 10 of each colour: three \textbf{1}s, two \textbf{2}s, \textbf{3}s, and \textbf{4}s, and finally a single \textbf{5}. The goal of the game is to play cards so as to form five consecutively ordered stacks, one for each colour, beginning with a card of rank \textbf{1} and ending with a card of rank \textbf{5}. What makes Hanabi special is that, unlike most card games, players can only see their partners' hands, and not their own.

\begin{figure}[h]
  \centering
  \includegraphics[width=\linewidth]{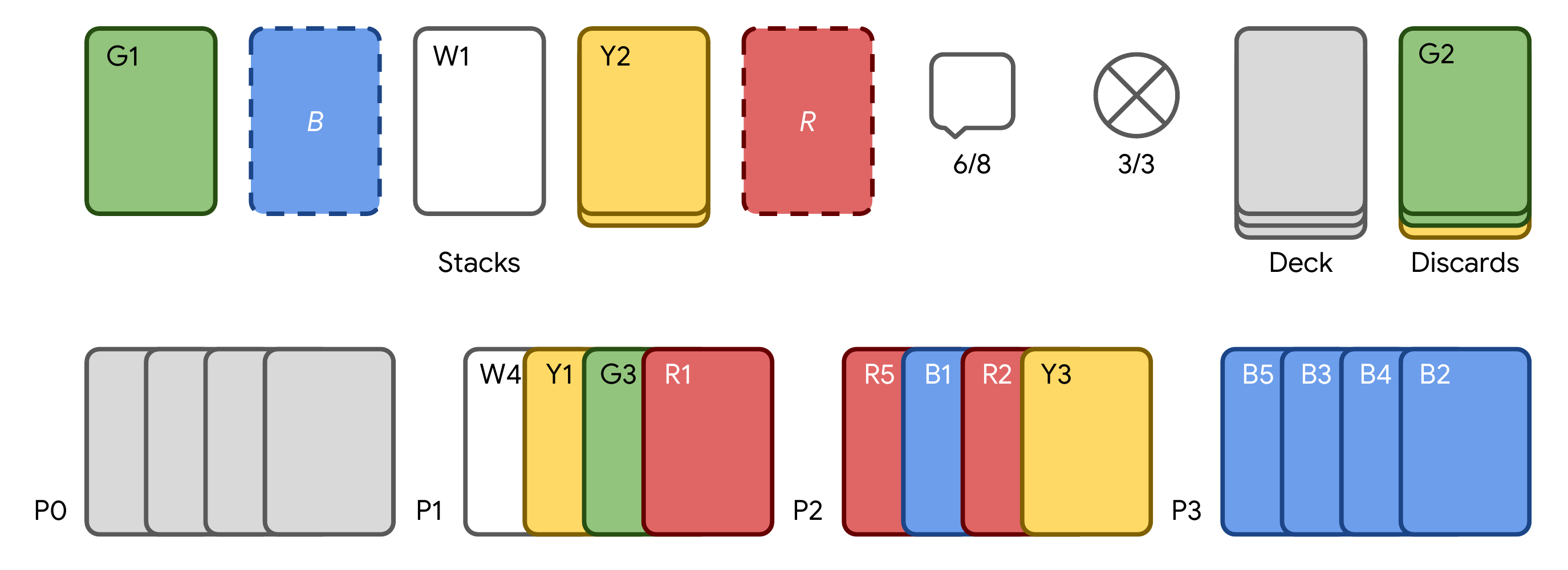}
  \caption{Example of a four player Hanabi game from the point of view of player 0. Player 1 acts after player 0 and so on.}
  \label{fig:example_game}
\end{figure}

Players take turns doing one of three actions: giving a hint, playing a card from their hand, or discarding a card. We call the player whose turn it is the \emph{active player}.

\noindent \textbf{Hints.} On their turn, the active player can give a hint to
any other player. A hint consists of choosing a rank or colour, and indicating
to another player all of their cards that match the given rank or colour. Only
ranks and colors that are present in the player's hand can be hinted for. For
example, in Figure~\ref{fig:example_game}, the active player may tell Player 2,
``Your first and third cards are \textbf{red}.'' or ``Your fourth card is a
\textbf{3}.'' To make the game interesting, hints are in limited supply. The game begins with the group owning eight information tokens, one of which is consumed every time a hint is given. If no information tokens remain, hints cannot be given and the player must instead play or discard.

\noindent \textbf{Discard.} Whenever fewer than eight information tokens remain, the active player can discard a card from their hand. The discarded card is placed face up (along with any unsuccessfully played cards), visible to all players. Discarding has two effects: the player draws a new card from the deck and an information token is recovered.

\noindent \textbf{Play.} Finally, the active player may pick a card
from their hand and attempt to play it. Playing a card is successful
if the card is the next in the sequence of its colour to be
played. For example, in Figure~\ref{fig:example_game}\ Player 2's action would be successful if they play their \textbf{yellow 3} or their \textbf{blue 1}; in the latter case forming the beginning of the blue stack. If the play is successful, the card is placed on top of the corresponding stack. When a stack is completed (the \textbf{5} is played) the players also receive a new information token (if they have fewer than eight). The player can play a card even if they know nothing about it; but if the play is unsuccessful, the card is discarded (without yielding an information token) and the group loses one life, possibly ending the game. In either circumstances, a new card is drawn from the deck.

\noindent \textbf{Game Over.} The game ends in one of three ways: either
because the group has successfully played cards to complete all five stacks,
when three lives have been lost, or after a player draws the last card from the
deck and every player has taken one final turn. If the game ends before three
lives are lost, the group scores one point for each card in each stack, for a
maximum of 25; otherwise, the score is 0.\footnote{Note that while scoring zero when the team runs out of lives agrees with the game's published rules, much of the prior research on Hanabi (discussed in Section~\ref{sec:related:hanabi}) scores this as the number of cards successfully played prior to failure.}

\subsection{Basic Strategy}
There are too few information tokens to provide complete information
(\emph{i.e.}, the rank and colour) for each of the 25 cards that can be played
through only the grounded information revealed by hints.~\footnote{The
deck contains 50 cards, 25 of which players aim to play.  When the deck runs
out the players will always hold at least 10 cards that cannot recover usable
information tokens, but each player could hold one of the 25 cards they want to
play. This means at most 17 cards can be discarded to recover information
tokens. Combined with the eight initial information tokens, and the four
recovered through completing colour stacks, this means players can hint at most
29 times during the game (and fewer times as the number of players
increases).}  While the quantity of information provided by a hint can be
improved by revealing information about multiple cards at once, the value of
information in Hanabi is very context dependent. To maximize the team's score
at the end of the game, hints need to be selected based on more than just the
quantity of information conveyed.  For example in
Figure~\ref{fig:example_game}, telling Player 3 that they hold four blue cards
reveals more information than telling Player 2 that they hold a single
rank-\textbf{1} card, but lower-ranked cards are more important early on, as
they can be played immediately. A typical game therefore begins by hinting to
players which cards are \textbf{1}s, after which those players play those
cards; this both ``unlocks'' the ability to play the same-colour \textbf{2}s
and makes the remaining \textbf{1}s of that colour useful for recovering
information tokens as players can discard the redundant cards.  

Players are
incentivized to avoid unsuccessful plays in two ways: first, losing all three
lives results in the game immediately ending with zero points; second, the card
itself is discarded. Generally speaking, discarding all cards of a given rank
and colour is a bad outcome, as it reduces the maximum achievable score.  For
example, in Figure~\ref{fig:example_game} both \textbf{green} \textbf{2}s have
been discarded, an effective loss of four points as no higher rank green cards
will ever be playable.  As a result, hinting to players that are at risk of
discarding the only remaining card of a given rank and colour is often
prioritized.  This is particularly common for rank-\textbf{5} cards since there
is only one of each colour and they often need to be held for a long time
before the card can successfully be played.

\subsection{Implicit Communication} 
While explicit communication in Hanabi is limited to the hint actions,
every action taken in Hanabi is observed by all players and can also implicitly
communicate information.  This implicit information is not conveyed through the
impact that an action has on the environment (\emph{i.e.}, what happens) but
through the very fact that a player decided to take this action
(\emph{i.e.}, why it happened).
This requires that players can reason over the actions that another player
would have taken in a number of different situations, essentially reasoning over the intent of the agent.
Human players often exploit such reasoning to convey more information through
their actions.  Consider the situation in Figure~\ref{fig:example_game} and
assume the active player (Player 0) knows nothing about their own cards, and so
they choose to hint to another player.  One option would be to tell Player 1 about
the \textbf{1}s in their hand.  However, that information is not particularly
actionable, as the \textbf{yellow 1} is not currently playable.  Instead, they
could tell Player 1 about the \textbf{red} card, which is a \textbf{1}.
Although Player 1 would not explicitly know the card is a \textbf{1}, and
therefore playable, they could infer that it is playable as there would be
little reason to tell them about it otherwise, especially when Player 2 has a
\textbf{blue 1} that would be beneficial to hint.  They may also infer that
because Player 0 chose to hint with the colour rather than the rank, that one
of their other cards is a non-playable \textbf{1}.

An even more effective, though also more sophisticated, tactic commonly
employed by humans is the so-called ``finesse'' move.  To perform the finesse
in this situation, Player 0 would tell Player 2 that they have a \textbf{2}.
By the same pragmatic reasoning as above, Player 2 could falsely infer that
their \textbf{red} \textbf{2} is the playable \textbf{white 2} (since both
\textbf{green 2}s were already discarded).  Player 1 can see Player 2's
\textbf{red 2} and realize that Player 2 will make this incorrect inference and
mistakenly play the card, leading Player 1 to question why Player 0 would have
chosen this seemingly irrational hint.  Even without established conventions,
players could reason about this hint assuming others are intending to
communicate useful information.  Consequently, the only rational explanation for
the choice is that Player 1 themselves must hold the \textbf{red 1} (in a
predictable position, such as the most recently drawn card) and is expected to
rescue the play.  Using this tactic, Player 0 can reveal enough information to
get two cards played using only a single information token.  There are many
other moves that
rely on this kind of reasoning about intent to convey useful information
(\emph{e.g.}, bluff, reverse finesse)~\cite{IraciHanabiConventions,ZamiellHanabiConventions}.  We will
use \emph{finesse} to broadly refer to this style of move.

%% file: core/5_the_benchmark.tex
\section{Hanabi: The Challenge}
\label{sec:benchmark}

We propose using Hanabi as a challenging benchmark problem for AI.
It is a multi-agent learning problem, unlike, for example,
the Arcade Learning Environment~\citep{bellemare13arcade}. It is also an
imperfect information game, where players have asymmetric knowledge about
the environment state, which makes the game more like poker than
chess, backgammon, or Go. The cooperative goal of Hanabi 
sets it further apart from all of these other challenge problems, 
which have players competing against each other. 
This combination of partial observability and
cooperative rewards creates unique challenges around the learning of
policies and communication. 
Unlike signalling games~\citep{lewis2008convention} the communication in Hanabi  does not use a separate channel, but rather mixes communication and environment actions. Finally, the resulting coordination and communication problem in Hanabi was designed to be challenging 
to human players. 

How humans play the game suggest two different challenges presented by
Hanabi.  The first, likely easier, challenge is to learn a fixed
policy for all players through \emph{self-play}. In this case, the
learning process is in control of all players, and the objective is to
maximise the expected utility of the resulting joint policy. This
represents the case where human players can pre-coordinate their play.
The second challenge, \emph{ad-hoc} team play, is learning to play
with a set of unknown partners, with only a few
games of interaction.

In the spirit of the proposed evaluation protocols~\citep{Mochado18} for the
Arcade Learning Environment, which discuss recommendations for learning in
Atari games, we will make some recommendations on how research should be
carried out under each challenge.  

\subsection{Open Source Environment}
\label{sec:open-environment}
To help promote consistent, comparable, and reproducible research
results, we have released an open source Hanabi reinforcement learning
environment called the \emph{Hanabi Learning
Environment}~\citep{HanabiLearningEnvironment}.
Written in Python and C++, the code
provides an interface similar to OpenAI Gym~\citep{OpenAIGym}. It
includes an environment state class which can generate observations
and rewards for an agent, and can be advanced by one step given agent
actions. An agent only needs to be able to generate an integer action,
given an observation bitstring.

The default agent observation in our Hanabi environment  includes
card knowledge from previous hint actions, which includes both positive and negative information (\emph{e.g.}, ``the first card is \textbf{red}'' also says all other cards are not \textbf{red}). This removes the memory
task from the challenge, but humans tend to find remembering cards to
be an uninteresting distraction, and the experimental results in
Section~\ref{sec:experiments} show that the game remains challenging
without requiring agents to learn how to remember card knowledge. For
researchers interested in memory, we provide the option to request a
minimal observation, which does not include this remembered card knowledge.

For debugging purposes, the code includes an environment for a small
game with only two colors, two cards per player, three information
tokens, and a single life token. There is also a very small game with
a single color.

\subsection{Challenge One: Self-Play Learning}
\label{sec:benchmark:self-play}
The self-play challenge is focused on finding a joint policy that
achieves a high expected score entirely through self-play learning.
A practical advantage of the Hanabi
benchmark is that the environment is extremely lightweight, both in terms of
memory and compute requirements, and fast (around 0.1ms per turn on a CPU). It
can therefore be used as a testbed for RL methods that require a large number
of samples without causing excessive compute requirements. However,
developing sample efficient algorithms is also an important goal for RL
algorithms in its own right. 
With this in mind we propose two different regimes for the Hanabi benchmark: 

\textbf{Sample limited regime (SL)}. In the \emph{sample limited
  regime}, we are interested in pushing the performance of
sample-efficient algorithms for learning to play Hanabi. To that
extent, we propose to limit the number of environment steps that the
agent can experience to be at most 100 million. Here \emph{environment
  steps} count the total number of turns taken during training. If the
current episode does not end at 100 million steps, then we let the
agent finish the episode before terminating training.  This regime is
similar to the evaluation scheme for Atari 2600 games proposed by
Machado and colleagues~\citep{Mochado18}. The 100 million step limit was chosen based on the learning curves of the Rainbow agent presented in Section~\ref{sec:experiments} to make sure that the current state-of-the-art agents can achieve a decent score in the given amount of time. 
Our intention with the sample limited regime is to highlight general
techniques that efficiently learn to play capably. Consequently, we
strongly encourage researchers not just to pursue techniques that
notably improve performance at the 100 million step horizon, but to
also demonstrate algorithms that achieve good performance with
dramatically fewer samples, closer to how human teams might experience
the game.

\textbf{Unlimited regime (UL)}. In the \emph{unlimited regime} there are no restrictions on the amount of time or compute. 
The unlimited regime describes research where the focus is on asymptotic performance, such as achieving high performance
using large-scale computation.
However, we encourage all work on Hanabi to include details about the compute requirements and run-time of their methods alongside the final results.

For every $k$-player game (where $k \in \{2,3,4,5\}$), we recommend the following details be reported. Here the \emph{best agent} is the training run with the highest average score under test conditions at the end of training, \emph{e.g.}, when disabling exploration and picking the greedy action.
\begin{itemize}
\item Training curves for all random number generator seeds, highlighting the best agent.
\item A histogram of game scores for the best agent and the percentage of perfect games.
\item The mean and standard deviation of the performance of the best agent, computed as an average performance across at least 1000 trials (\emph{i.e.}, full games). In the future, as performance increments become smaller this number should be increased to allow for significant results.
\end{itemize}

As we will show in Section~\ref{sec:experiments}, Hanabi is
difficult for current learning algorithms. Even when using a large amount of
data and computation time (UL regime), learning agents have trouble approaching
the performance of hand-crafted rules in four player games, and
fall far short of such rules for three and five players.  

\subsection{Challenge Two: Ad-hoc Teams}
\label{sec:benchmark:ad-hoc}

The second challenge Hanabi poses is that of ad-hoc team play.
The ultimate goal is agents that are capable of playing with other agents or
even human players. For this, a policy which achieves a high score in self-play is of little
use if it must be followed exactly by teammates. Good strategies are not
unique, and a robust player must learn to recognize intent in other agents' actions and adapt to a wide
range of possible strategies.

We propose to evaluate ad-hoc team performance by measuring an
agent's ability to play with a wide range of teammates it has never encountered
before.   This is measured via the score achieved by the agent when it
is paired with teammates chosen from a held-out pool of agents.~\footnote{For two-player games forming a team is straightforward: we pair the evaluated
agent with a random player from the pool and randomly permute their order in the
team. For three to five-player games, the presence of more than one player from the pool is a
potential confounding factor: it could be that the team fails because of the
interaction between these players and through no fault of the evaluated agent.
We therefore recommend to limit teams to two unique players --- the evaluated
agent, in one seat, and one agent from the pool --- which is
replicated in the remaining seats.}
The composition of
that pool should be such that the players exhibit diverse strategies, which can
be hard-coded or learned by self-play.

We recommend the evaluated agents be given ten random self-play games of its
ad-hoc teammates prior to play.  While other alternatives may be more
challenging (e.g., ten ``warm-up'' games, or average performance in
the first ten games with no prior information), this focuses the
challenge on an agent's ability to recognize intent in other agents'
behaviour, as they can observe examples of the intended properly
coordinated behaviour.  

We recommend that the mean and standard deviation of the performance be
reported across at least 1000 trials for each hold-out team.  Specifically,
each trial for a particular hold-out team should be evaluated by giving the
agent a set of ten self-play games for the team, followed by the agent playing
a \emph{single} game in place of a player from the hold-out team in a random
position, and finally resetting the agent so it does not retain memory across
trials.  The 1000 trials should also involve at least 100 different random sets
of self-play games provided to the agent.  These results should be reported
along with mean and standard deviations of the performance of the hold-out
teams in self-play (as a baseline comparison).  We further recommend
crosstables of the hold-out teams' performance when paired with each other as a
method of assessing the diversity of the pool (e.g., see
Figure~\ref{fig:firebeast-results-mixing-selfplay-strategies}).

In the future we expect to see canonical agent pools of pre-trained or
hard-coded self-play agents be made available for training and
hold-out sets to allow for consistent comparisons.

%% file: core/6_empirical_evidence.tex
\section{Hanabi: State of the Art}
\label{sec:experiments}


Hanabi presents interesting multi-agent learning challenges for both
learning a good self-play policy and adapting to an ad-hoc team of players.  In this
section, we provide empirical evidence that both tasks are challenging for
state-of-the-art learning algorithms, even with an abundance of computational
resources.

As in many domains, we would ultimately like to contrast current
machine learning techniques with human performance. Unfortunately, empirically evaluating human
performance is difficult due to a variety of factors: identifying and
recruiting players, gathering a statistically meaningful quantity of data
without having players suffer from fatigue, and mitigating other potentially
confounding issues.  For instance, from informally looking at online Hanabi
game rooms, human players often abandon games part way through when they are
unable to achieve perfect scores, causing bias in the data.  Moreover, humans
are also able to exploit other non-verbal cues such as eye
movements~\citep{GottwaldEM18} and even the time it takes for their partners to
make decisions. Instead of human-level ``self-play'', which one might consider as the performance of an established team of humans, we contrast performance with hand-coded ``bots'' that make use of human-style conventions.

In the self-play setting, we examine the performance of two modern
multi-agent reinforcement learning algorithms using deep learning for
function approximation. We contrast these methods with a few of the
best known hand-coded Hanabi ``bots'', and show that the learning
agents fall somewhere between being competitive with hand-coded rules and
being significantly outperformed by them. This failure is not just a
matter of poor data efficiency: even with an intentionally
unreasonable amount of experience and computation, the learning agents are
outperformed by hand-coded rules.

We also show self-play results for a recent reinforcement learning
algorithm that was designed to tackle the joint learning problem posed
by Hanabi, explicitly reasoning about both public information and what
other agents might have privately observed. This agent was only run
for two player Hanabi, but achieved the best reported self-play
performance in this case.

Finally, in the ad-hoc setting we show the performance for the two
modern reinforcement learning algorithms, using agents from multiple
independent runs. In this case, combinations of different learning
agents only score slightly more than zero points.

We begin our empirical analysis with an overview of the different
learning agents and rule-based bots used in our experiments.

\subsection{Learning Agents}
\noindent \textbf {Actor-Critic-Hanabi-Agent.}
The family of asynchronous advantage actor-critic
algorithms~\citep{Mnih16asynchronous} demonstrates stability, scalability and
good performance on a range of single-agent tasks, including the suite of games
from the Arcade Learning Environment~\citep{bellemare13arcade}, the TORCS driving
simulator~\citep{Wymann2015TORCS}, and 3D first-person environments~\citep{Beattie16DMLab}.
In the original implementation, the policy is represented by a deep neural network,
which also learns a value function to act as a baseline for variance reduction.
Experience is accrued in parallel by several copies of the agent running in
different instantiations of the environment. Learning gradients are passed back
to a centralized server which holds the parameters for a deep neural network.

Since the environment instantiations and the server interact asynchronously, there
is a potential for the learning gradients to become stale, which impacts negatively
on performance.  ACHA uses the Importance Weighted Actor-Learner
variant to address the stale gradient
problem~\citep{Espeholt18IMPALA} by adjusting the stale off-policy
updates using the V-trace algorithm. 
The variant has been successfully applied to the multi-agent task of Capture-the-Flag,
achieving human-level performance~\citep{Jaderberg18CTF,Jaderberg19CTF}.  ACHA also  incorporates
population-based training~\citep{Jaderberg17PBT}, providing automatic hyperparameter optimization.

For our experiments, ACHA was run with a population size of $30$ to  $50$ per
run, $100$ actors generating experience in parallel, and hyperparameter
evolution over the learning rate and entropy regularisation weight. 
ACHA also used parameter-sharing across the different players in combination with an
agent-specific ID that is part of the input. Parameter sharing is a standard method which increases learning speed,
while the agent-specific ID allows for some level of specialisation between agents.
Our neural
architecture consisted of the following. All observations were first processed by
an MLP with a single 256-unit hidden layer and ReLU activations, then fed into a
2-layer LSTM with 256 units in each layer. The policy $\pi$ was a softmax readout
of the LSTM output, and the baseline was a learned linear readout of the LSTM output. We refer to this method as the Actor-Critic-Hanabi-Agent (\acha{}).
To demonstrate what is possible in the unlimited regime, we trained ACHA
agents for 20 billion steps. We estimate the computation took 100 CPU years for
a population size of 30, however this is likely an overestimate as CPU usage was
not saturated and jobs were able to be preempted by the cluster’s scheduler.


\noindent \textbf {Rainbow-Agent.} Rainbow~\citep{Hessel17Rainbow} is a state of the art agent architecture for deep RL on the Arcade Learning Environment. It combines some of the key
innovations that have been made to Deep Q-Networks (DQN)~\citep{mnih15human} over the last few years, resulting in a
learning algorithm that is both sample efficient and achieves high rewards at
convergence. In our benchmark we use a multi-agent version of Rainbow, based on
the Dopamine framework~\citep{castro18dopamine}. In our code 
the agents controlling the different players share parameters. Our Rainbow agent is feedforward and does not use any observation stacking outside of the last action, which is included in the current observation.

Our Rainbow agent uses a 2-layer MLP of $512$ hidden units each to predict
value distributions using distributional reinforcement
learning~\citep{bellemare17distributional}. Our batch size, \emph{i.e.}, the
number of experience tuples sampled from the replay buffer per update,  is
$32$, and we anneal the $\epsilon$ of our $\epsilon$-greedy policy to 0 over
the course of the first $1000$ training steps. We use a discount factor,
$\gamma$, of $0.99$ and apply prioritized sampling~\citep{schaul16prioritized} for sampling from the replay buffer. Finally, our value distributions are approximated as a discrete distribution over $51$ uniformly spaced atoms. 
Training a Rainbow agent for the sample limited regime of 100 million
steps took approximately seven days using an NVIDIA V100 GPU.

\noindent \textbf {BAD-Agent~\citep{Foerster18BAD}. }
For the two player self-play setting we also include the results of the Bayesian Action
Decoder since it constitutes state-of-the-art for the two-player unlimited regime. 
 Rather than relying on implicit belief representations such as RNNs, the Bayesian Action Decoder (BAD) uses a Bayesian belief update that directly conditions on the current policy of the acting agent. 
 In BAD all agents track a public belief, which includes everything that is common knowledge about the cards, including the posterior that is induced from observing the actions different agents take.  
 BAD also explores in the space of deterministic  policies, which ensures informative posteriors while also allowing for randomness required to explore.
Further
details for the BAD agent are provided in~\citep{Foerster18BAD}. 

\subsection{Rule-Based Approaches}
For benchmarking we provide results of a number of independently
implemented rule-based strategies.  Unlike the previous learning agents,
which learn a policy without explicitly encoding conventions for behaviour,
these rule-based strategies directly encode conventions through their rules.
These bots provide examples of the quality of play that can be achieved in
Hanabi.  We focus our benchmarking on the following rule-based strategies because they outperform other prior works on Hanabi (which we discuss in Section~\ref{sec:related:hanabi}), most of which also exploit various prespecified rules in some manner.

\label{sec:experiments_rulebased}
\noindent \textbf {SmartBot~\citep{SmartBot}.} SmartBot is a
rule-based agent that tracks
the publicly known information about each player's cards. Tracking public
knowledge allows SmartBot to reason about what other players may do, and what
additional knowledge it gains from its specific view of the game. Among other
things, this enables SmartBot to play/discard cards that its partners
do not 
know that it knows are safe to play/discard, thereby preventing partners from
wasting a hint to signal as much.  However, this tracking assumes all other
players are using SmartBot's policy. When this assumption does not hold, as
in the ad-hoc team setting, SmartBot can fall into false or impossible beliefs.
For example, SmartBot can believe one of its cards has no valid value as all
possible cards are inconsistent with the observed play according to SmartBot's
convention.  Finally, note that SmartBot has a parameter specifying if it
should attempt uncertain plays that may cost a life.  Risking lives increases
the frequency of perfect games while reducing average score, except in two
player games where it is better on both criteria.  Our SmartBot results only
risks lives in the two player setting.

\noindent \textbf {HatBot~\citep{Cox15}  and WTFWThat~\citep{wtfwt}.} HatBot uses a technique often
seen in coding
theory and ``hat puzzles''. When giving hints, HatBot uses a predefined
protocol to determine a recommended action for all other players (\emph{i.e.},
play or discard for one of a player's cards). This joint recommendation is then
encoded by summing the indices for the individual recommendations and using
modular arithmetic. The encoded joint recommendation is mapped to different
hints that HatBot could make, specifically, whether it reveals the color or
rank of a card for each other player. Since each player can view everyone's
cards but their own, they can reconstruct the action recommended to them by
figuring out what would have been recommended to the other players based on
HatBot's convention, which HatBot assumes they know and use. Although this
convention is not very intuitive for human players, it would still be
possible for humans to learn and follow. Cox and colleagues also introduce an ``information
strategy'' using a similar encoding mechanism to directly convey information
about each player's cards (as opposed to a recommended action), however it
requires additional bookkeeping that makes it impractical for humans to use.
As originally proposed, both the recommendation and information strategies were
tailored for playing 5-player Hanabi. However, a variant of the information
strategy, called \emph{WTFWThat}~\citep{wtfwt}, can play two through five
players.  These benchmarks more provide a lower-bound for optimal play
than a baseline suggestive of human performance.

\noindent \textbf {FireFlower~\citep{FireFlower}.} FireFlower implements a set
of human-style conventions (detailed in \ref{app:fireflower}). The bot
keeps track of both private and common
knowledge, including properties of cards that are implied by the common
knowledge of what the conventions entail. Using this, FireFlower performs a
2-ply search over all possible actions with a modelled probability distribution over what its partner will do in response, and chooses the action that maximizes the expected value of an evaluation function. The evaluation function takes into account the physical state of the game as well as the belief state. For example, if there is a card in the partner's hand that is commonly known (due to inference from earlier actions) to be likely to be playable, then the evaluation function's output will be much higher if it is indeed playable than if it is not. 
FireFlower uses a few additional conventions for three and four players,
but avoids the hat-like strategies in favour of conventions that potentially
allow it to partner more naturally with humans. According to Fireflower's creator, it is designed with a focus on maximising the win probability, rather than average score. 

\subsection{Experimental Results: Self-Play}
\label{sec:experiments:self-play}
Table~\ref{tab:results_summary} shows the experimental results of the
baseline agents and state-of-the-art learning algorithms with each
number of players.  
\begin{table}[h]
{\scriptsize
\begin{center}
 \begin{tabular}{||c|c| c c c c||} 
 \hline
 Regime & Agent & 2P & 3P & 4P & 5P \\ [0.5ex] 
 \hline\hline
 -- & SmartBot &  22.99 (0.00) & 23.12 (0.00) & 22.19 (0.00) & 20.25 (0.00) \\
    &          &  29.6\% & 13.8\%  & 2.076\%  & 0.0043\% \\
 \hline

 -- & FireFlower &  22.56 (0.00) & 21.05 (0.01)  & 21.78 (0.01)  & - \\
    &        &  52.6\%         & 40.2\% & 26.5\% &   \\
               \hline
 -- & HatBot & -- & --  & --  &  22.56 (0.06) \\
    &        &    &     &    & 14.7 \%   \\   
       \hline
 -- & WTFWThat & 19.45 (0.03) & \textbf{24.20} (0.01) & \textbf{24.83} (0.01) & \textbf{24.89} (0.00)  \\
    &                         & 0.28\%  &  \textbf{49.1}\%  & \textbf{87.2}\% & \textbf{91.5}\%  \\   
      \hline
 \hline
SL & Rainbow & 20.64 (0.11) & 18.71 (0.10) & 18.00 (0.09)  & 15.26 (0.09) \\
 && 2.5 \% &  0.2\% & 0 \% & 0 \% \\
       \hline
UL & \acha & 22.73 (0.12) & 20.24 (0.15) & 21.57 (0.12) & 16.80 (0.13) \\
   &    & 15.1\%        & 1.1\%    & 2.4\%    & 0\%      \\
  \hline
  UL & BAD & \textbf{ 23.92 } (0.01) & --  & --  & -- \\
    &   & \textbf{ 58.6}\%    &     &     &    \\
 \hline
\end{tabular}
\end{center}
   \caption{Shown are the results for the three learning agents, Rainbow, \acha{} and BAD, compared to the rule-based agents, SmartBot, FireFlower and HatBot, for different numbers of players in \emph{self-play}. For each algorithm and number of players we show mean performance of the best agent followed by (standard error of the mean) and percentage of perfect (\emph{i.e.}, 25 point) games. Error of the mean differs based on different number of evaluation games.}
     \label{tab:results_summary}
}
\end{table}
First, note that neither the Rainbow nor \acha{} agent reaches the
performance of the best hand-coded agent in the two-player setting
(SmartBot), and neither learning agent reaches the performance of any
of the hand-coded agents with more than two players. There is a large
performance gap in what is possible (as demonstrated by the hand-coded
agents) and what state-of-the-art learning algorithms achieve.  Even
rule-based strategies that codify more human-like conventions achieve scores
higher than the learning algorithms, particularly in the three and five player
setting. Experienced human teams are generally considered stronger than such
bots, suggesting the gap between these learning agents and superhuman
performance in self-play is even larger still.

The final learning agent, BAD, only reports results for the two player
setting, but achieves the best reported performance in that setting.
Because BAD was only tested on two player Hanabi and a small synthetic
game, we can not make a strong claim that BAD is an algorithmic
improvement for a general class of games, rather than just being good
at Hanabi. However, we find it suggestive that BAD's tracking of all agents' beliefs leads to a marked improvement in performance
for an agent learning to play.

Notice that both BAD and the \acha{} agent require vast amounts of training data to establish good performance. As such, the self-play challenge, even for two player Hanabi, is fertile ground for further innovation. Furthermore, theory-of-mind-inspired methods still need to be scaled to settings with more than two players.

\begin{figure}[htp]
  \centering
  \vskip -4pt
  \begin{subfigure}[h]{0.405\linewidth}
    \includegraphics[width=\linewidth]{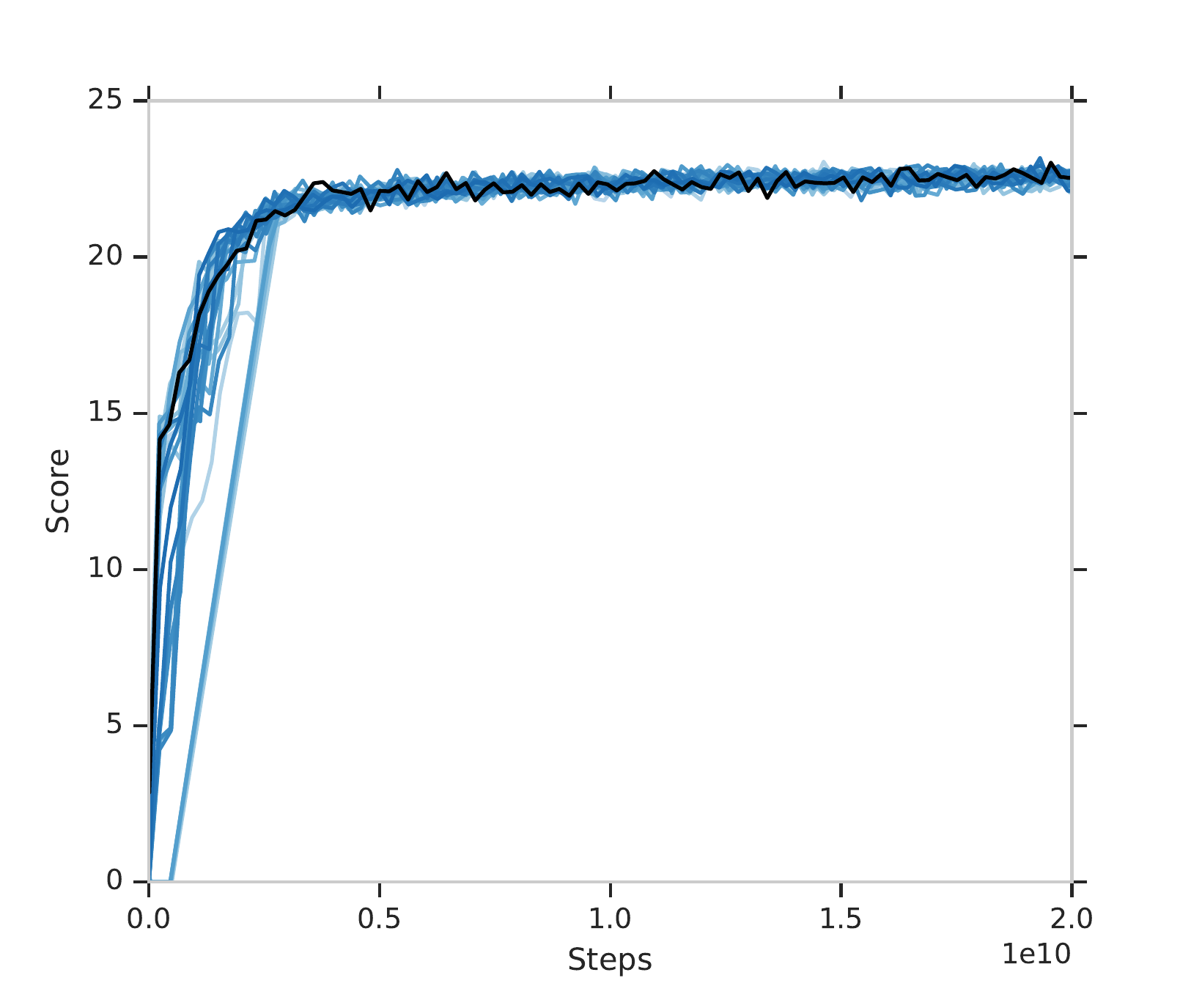}
  \end{subfigure}
  \begin{subfigure}[h]{0.405\linewidth}
    \includegraphics[width=\linewidth]{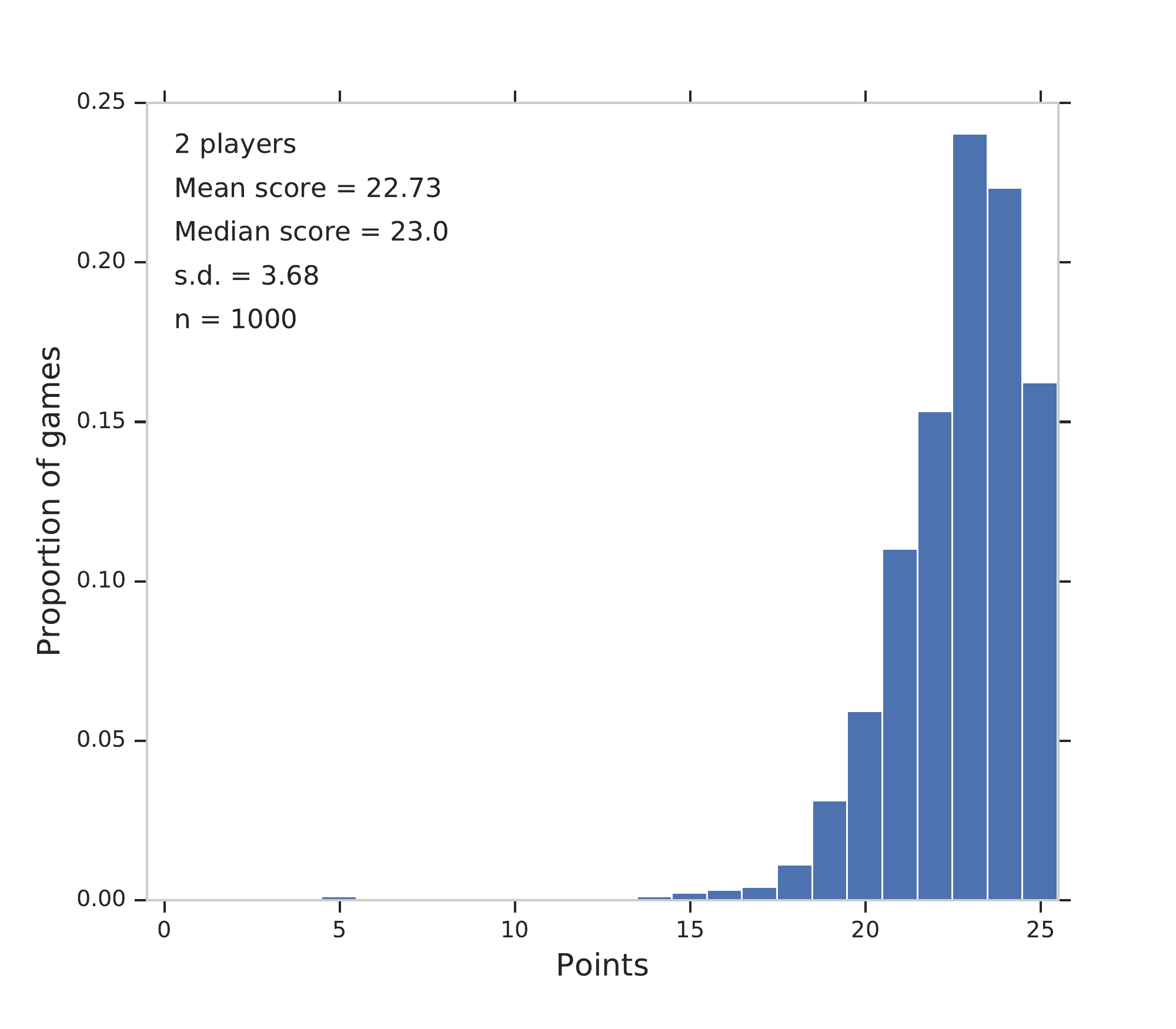}
  \end{subfigure}
  \vskip -8pt
  \begin{subfigure}[h]{0.405\linewidth}
    \includegraphics[width=\linewidth]{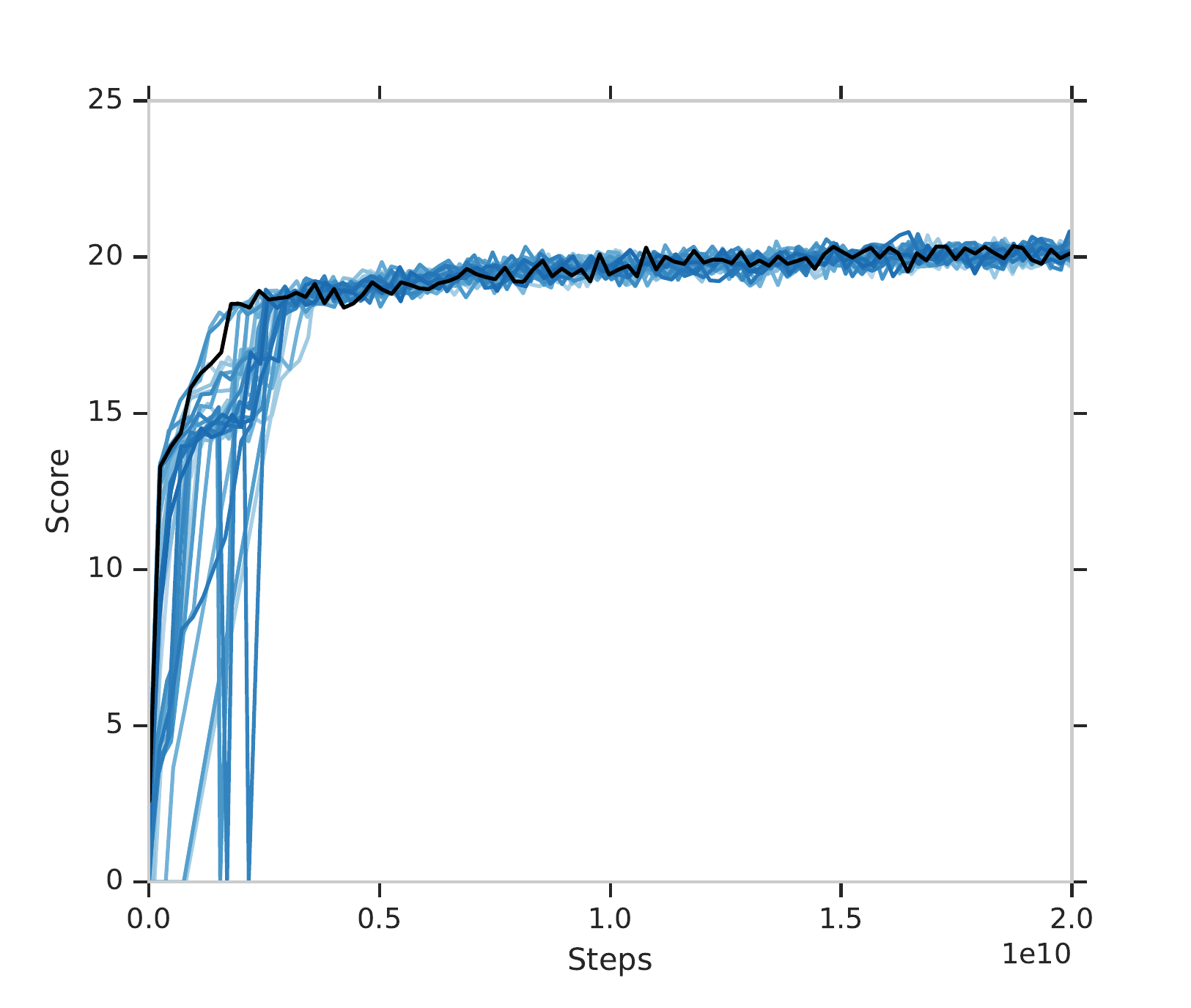}
  \end{subfigure}
  \begin{subfigure}[h]{0.405\linewidth}
    \includegraphics[width=\linewidth]{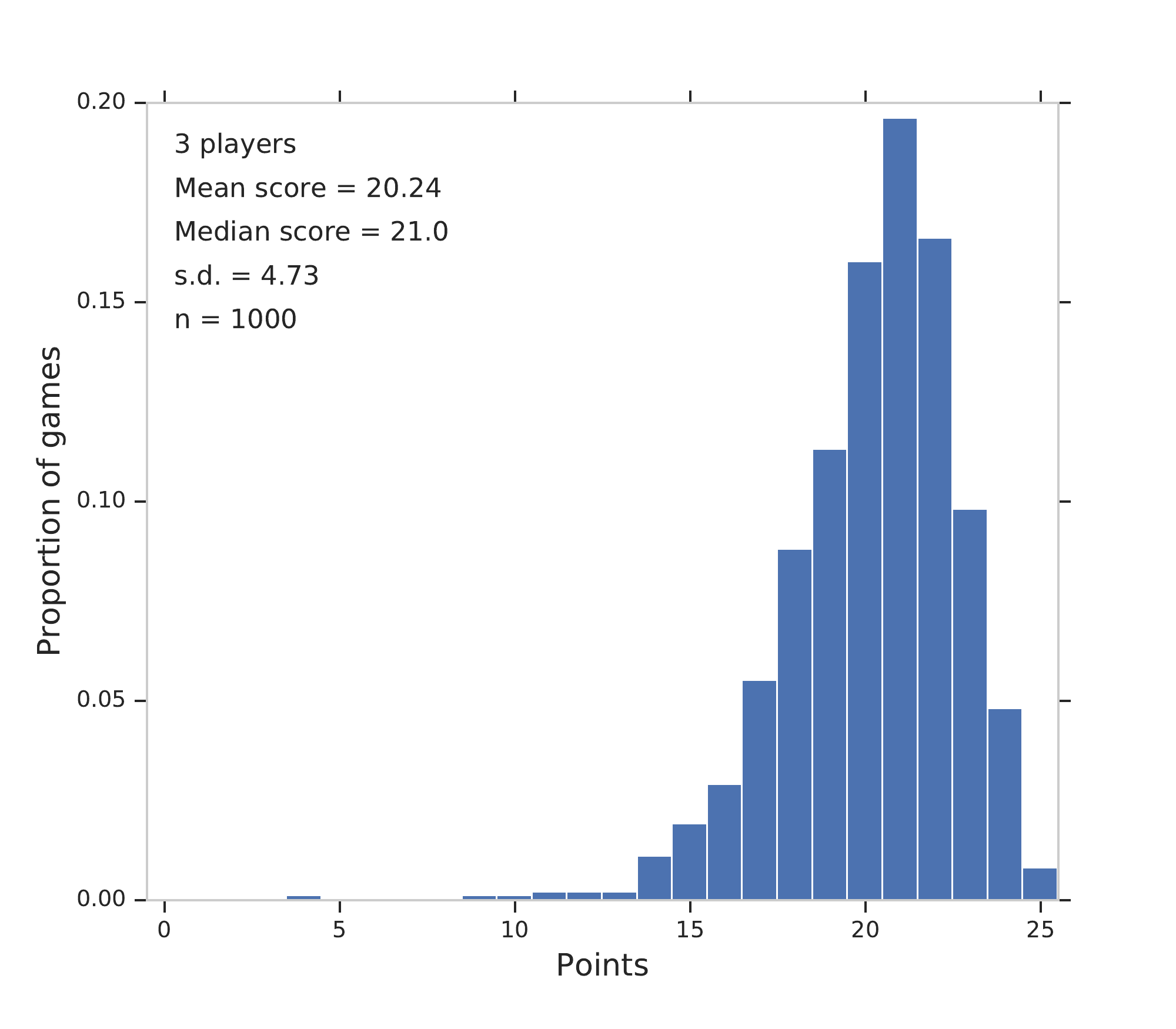}
  \end{subfigure}
  \vskip -8pt
  \begin{subfigure}[h]{0.405\linewidth}
    \includegraphics[width=\linewidth]{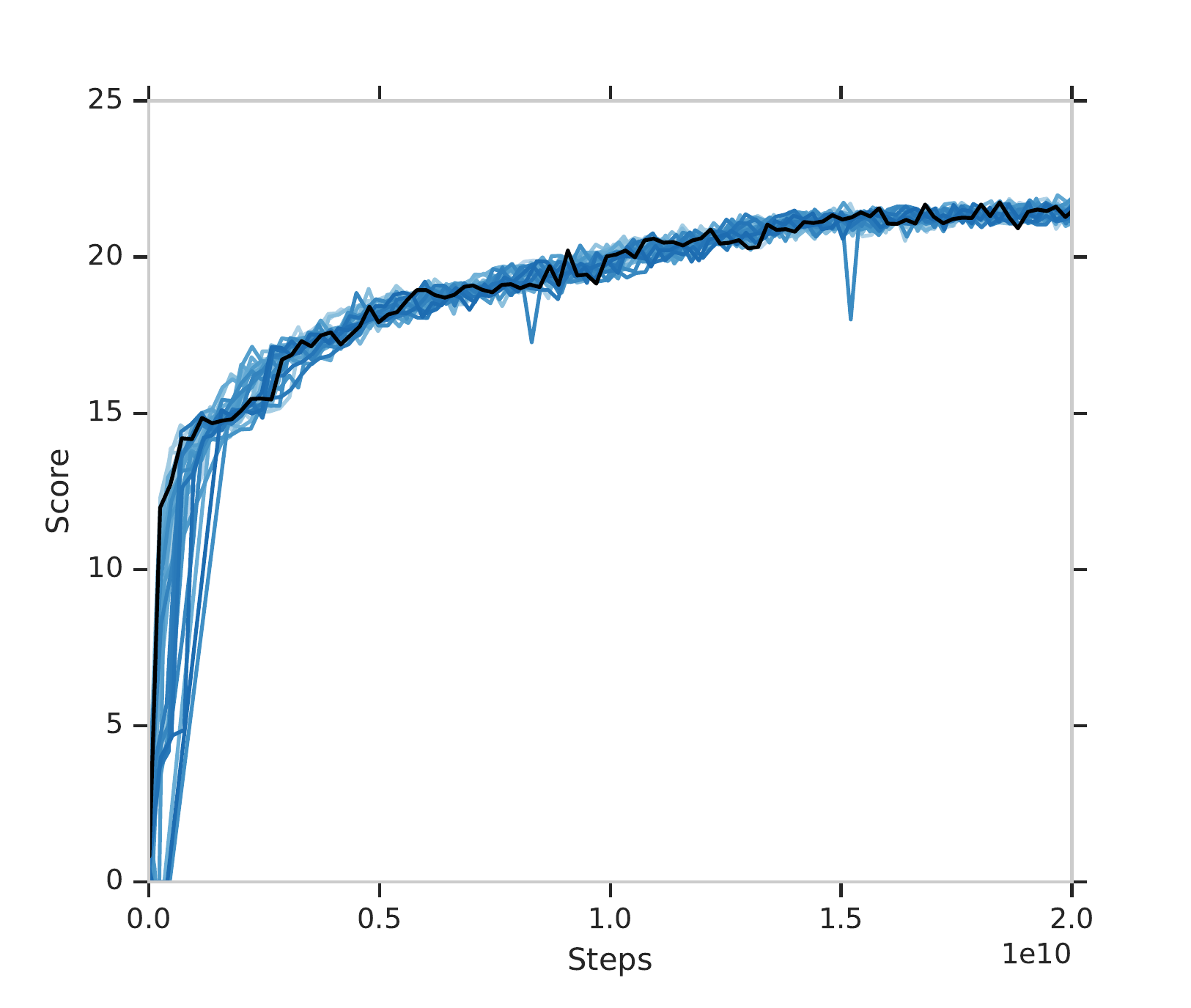}
  \end{subfigure}
  \begin{subfigure}[h]{0.405\linewidth}
    \includegraphics[width=\linewidth]{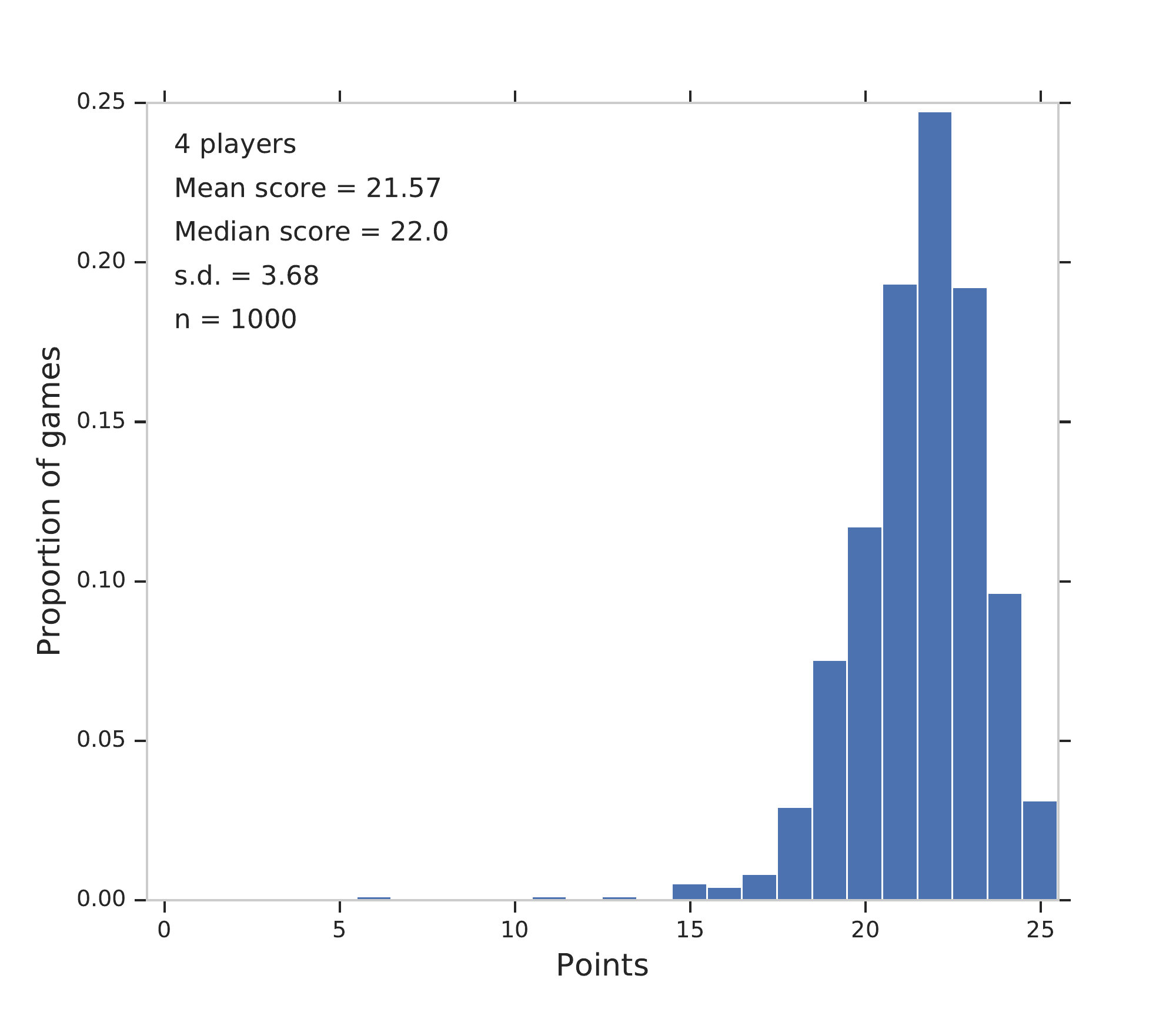}
  \end{subfigure}
  \vskip -8pt
  \begin{subfigure}[h]{0.405\linewidth}
    \includegraphics[width=\linewidth]{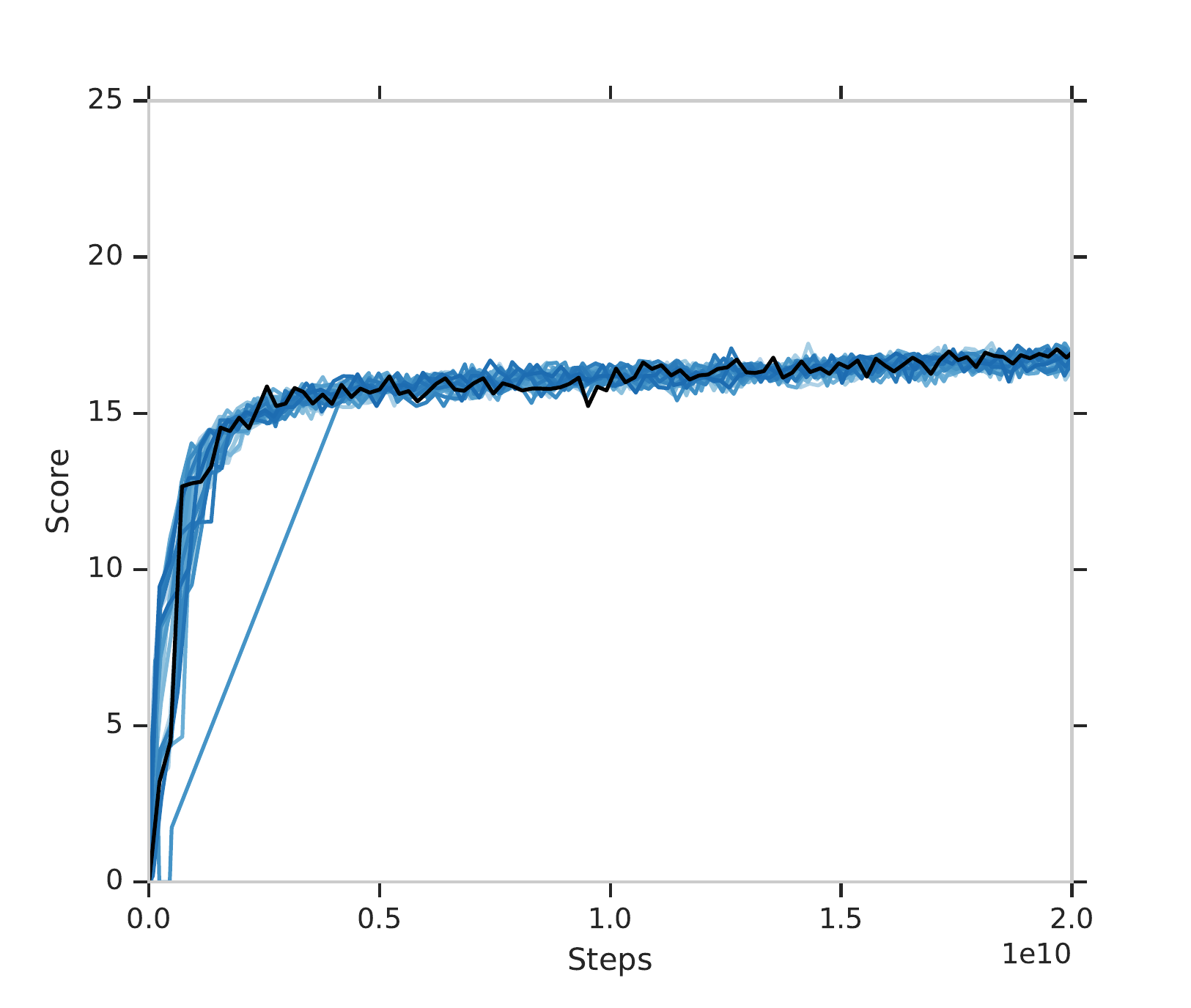}
  \end{subfigure}
  \begin{subfigure}[h]{0.405\linewidth}
    \includegraphics[width=\linewidth]{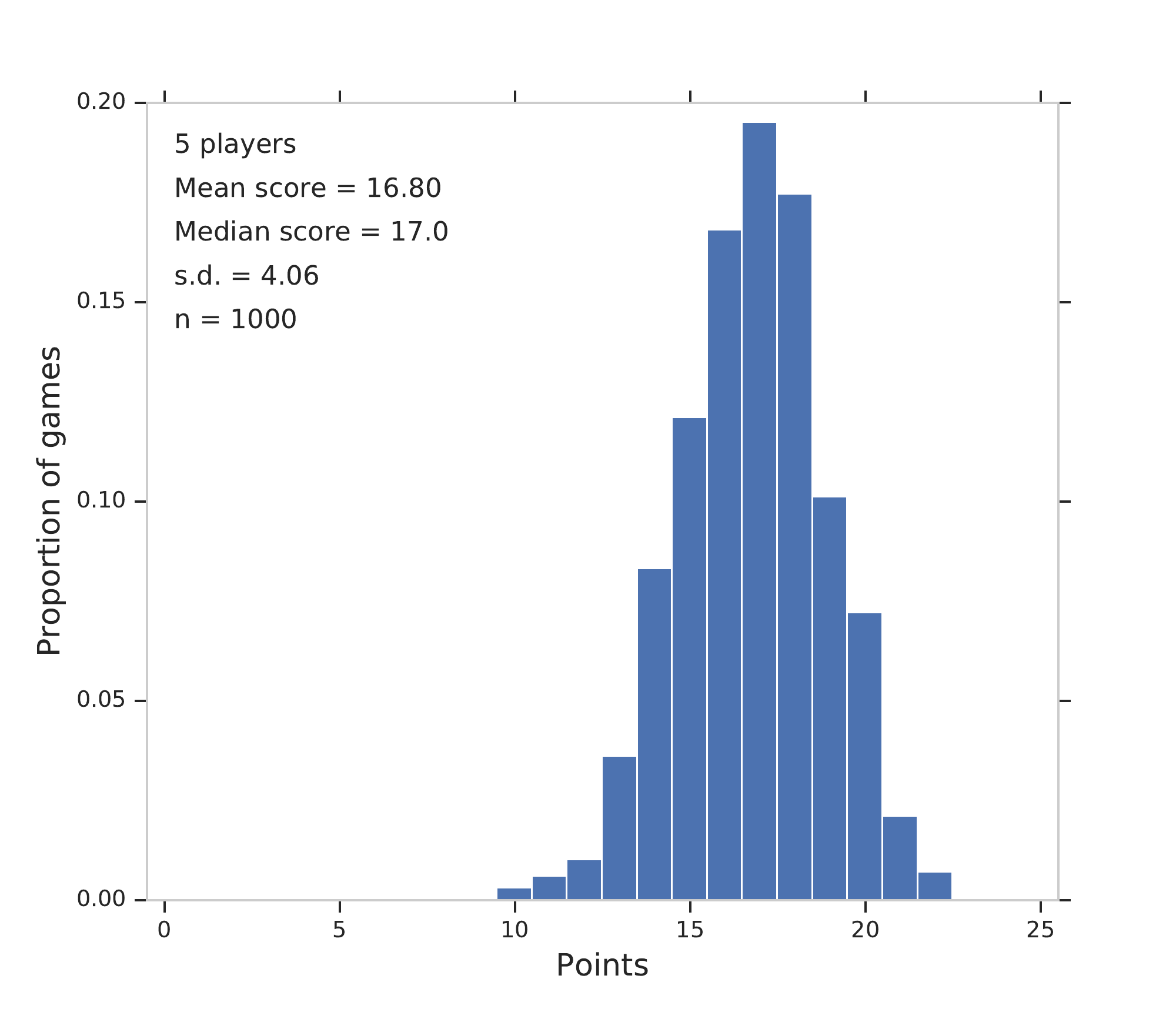}
  \end{subfigure}
  \caption{\acha{} results for two to five players, from top to bottom
respectively. Performance curves (left) are training-time results from the
current soft-max policy. Average scores and distributions (right) are test-time
results from 1000 episodes generated using the greedy policy from the agent
with the best training score.}
  \label{fig:firebeast-results}
\end{figure}

Comparing the more traditional RL approaches, the \acha{} agent in the
unlimited regime (using over 20 billion steps of experience for each
learner in the population) achieved higher scores than Rainbow (using 100
million steps of experience) across all number of players.  This quite
naturally may be due to \acha{} using more training experience,
but may also be due to Rainbow's feed-forward network architecture
with no history of past actions possibly making it harder to learn
multi-step conventions.  Both agents saw a decline in performance as
the number of agents increased with Rainbow's more gradual, while
\acha{} saw a precipitous drop in performance with five players.



\begin{figure}[h]
  \centering
  \begin{subfigure}[h]{0.42\linewidth}
    \includegraphics[width=\linewidth]{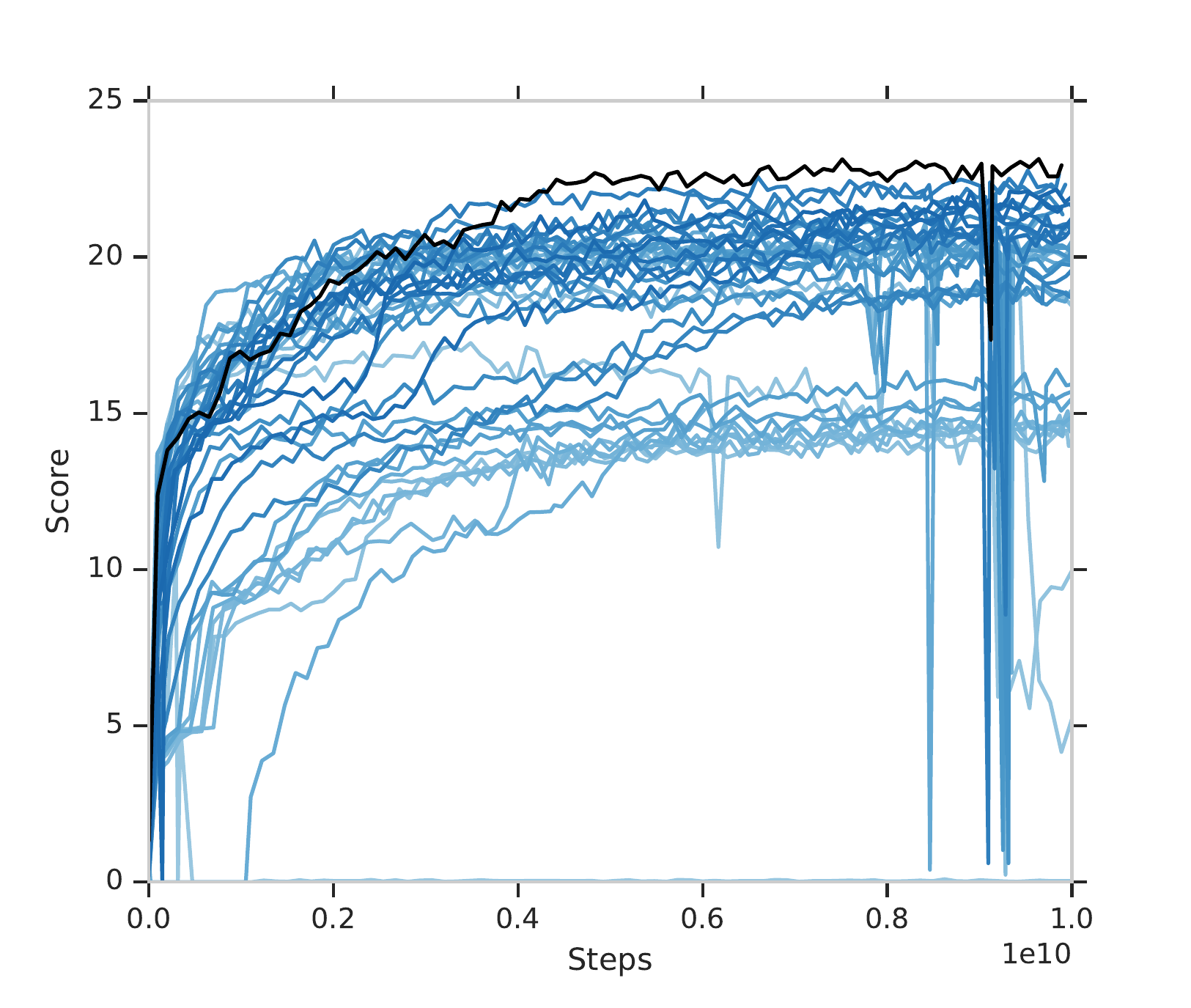}
  \end{subfigure}
  \begin{subfigure}[h]{0.42\linewidth}
    \includegraphics[width=\linewidth]{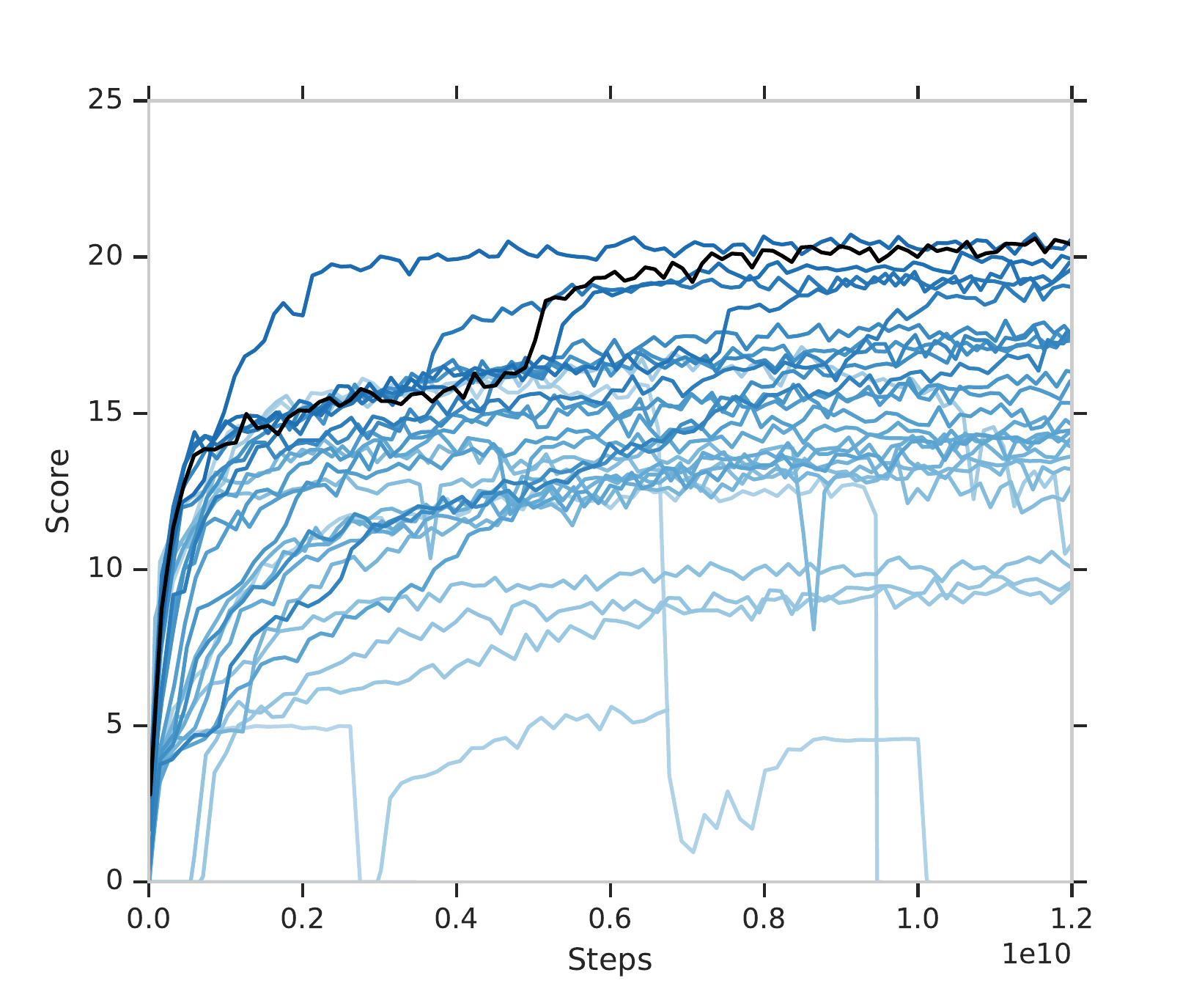}
  \end{subfigure}
  \caption{\acha{} results without evolution, using 50 independent runs
    for two players (left), and 30 independent runs for four players (right).}
  \label{fig:firebeast-results-no-evolution}
\end{figure}

Figure~\ref{fig:firebeast-results} shows
training curves for one run of \acha{} showing the performance of the multiple ``agents''  within
the population.  Note that these curves are linked together through
parameter evolution and are not independent. In all but the four
player setting, it appears \acha{} has found a local minimum in
policy space and it is unable to escape even with more training.
However, parameter evolution is hiding the true extent of the local
minima problem.
Figure~\ref{fig:firebeast-results-no-evolution} gives \acha{} 
training curves for two and four player games when evolution is
disabled, so each curve is an independent self-play trial, using the
same fixed hyper-parameters found as the best in the earlier
experiment.  Here we can see that independent learning trials find a
wide array of different local minima, most of which are difficult to escape.
For example, in the two player setting, roughly a third of the
independent agents are below 15 points and appear to no longer be
improving.

We further find that even \acha{} runs with similar final performance can learn
very different conventions. For example, one agent uses color hints to indicate
that the 4th card of the teammate can likely be discarded, while
another agent uses the different color hints to indicate which card of
the teammate can be played. Different agents also use the rank-hints
to indicate playability of cards in different slots.  Details
examining specific examples of learned policies are in \ref{app:conditionalprob}.

\begin{figure}[htp]
  \centering
  \vskip -16pt
  \begin{subfigure}[h]{0.405\linewidth}
    \includegraphics[width=\linewidth]{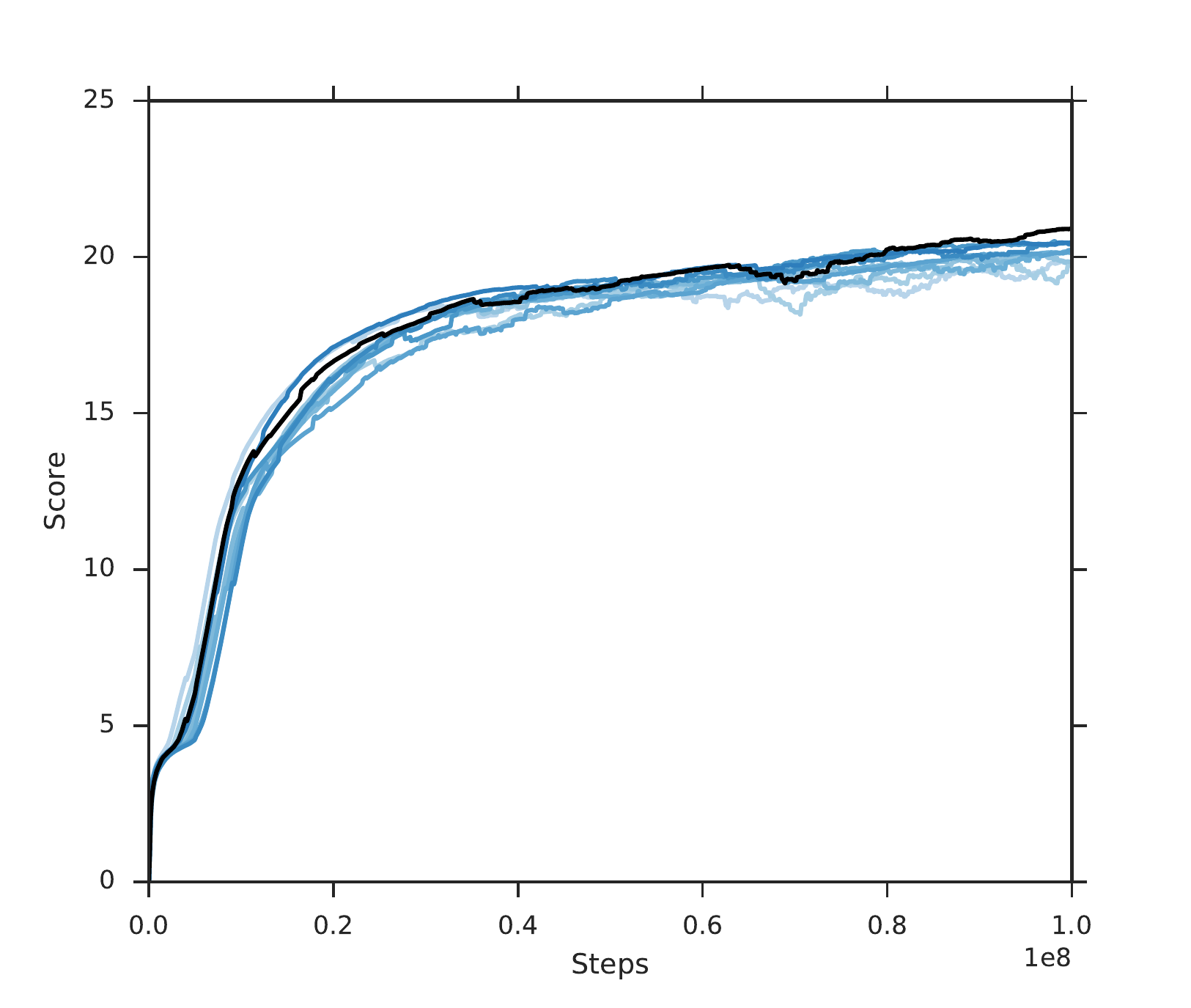}
  \end{subfigure}
  \begin{subfigure}[h]{0.405\linewidth}
    \includegraphics[width=\linewidth]{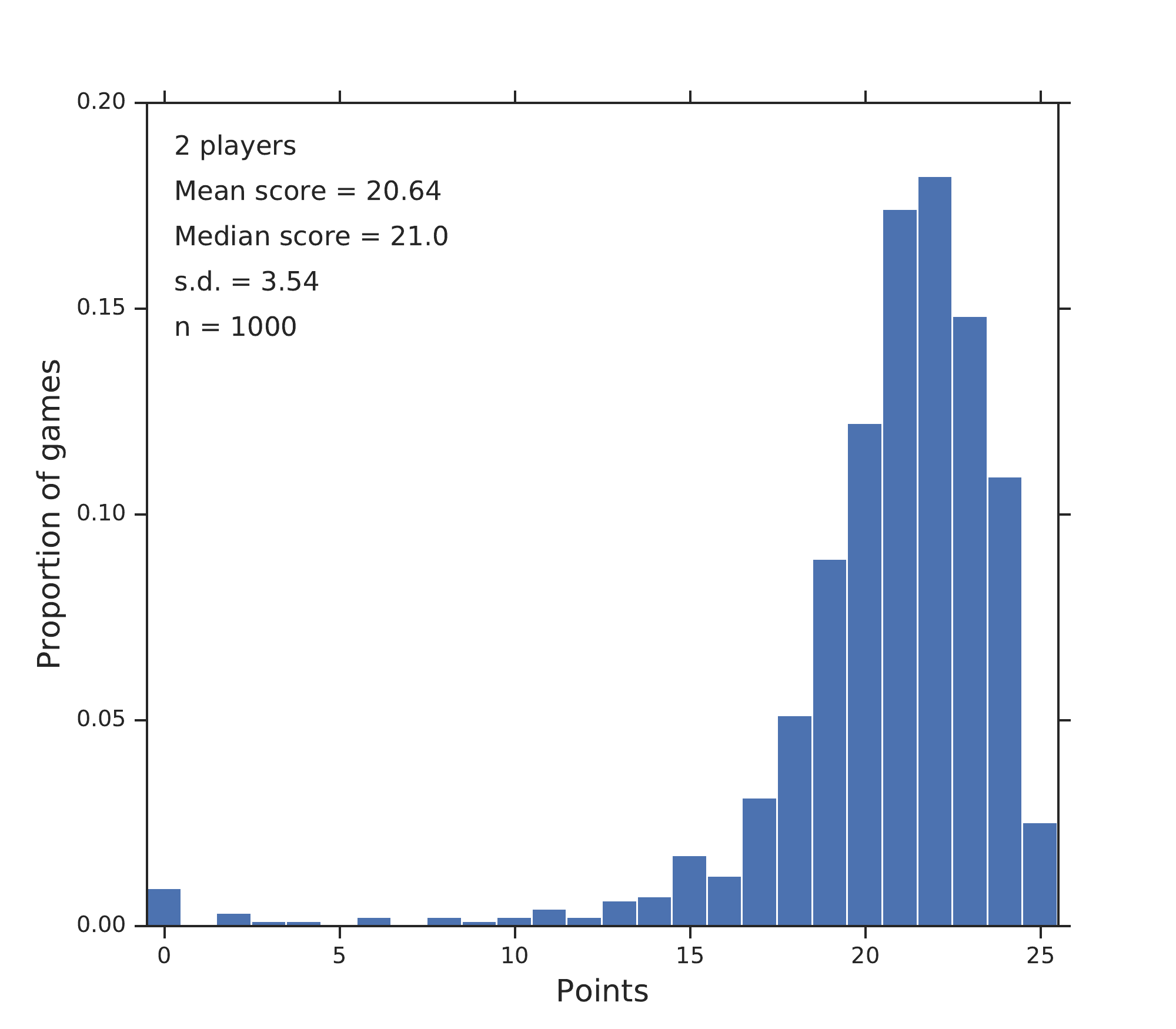}
  \end{subfigure}
  \vskip -4pt
  \begin{subfigure}[h]{0.405\linewidth}
    \includegraphics[width=\linewidth]{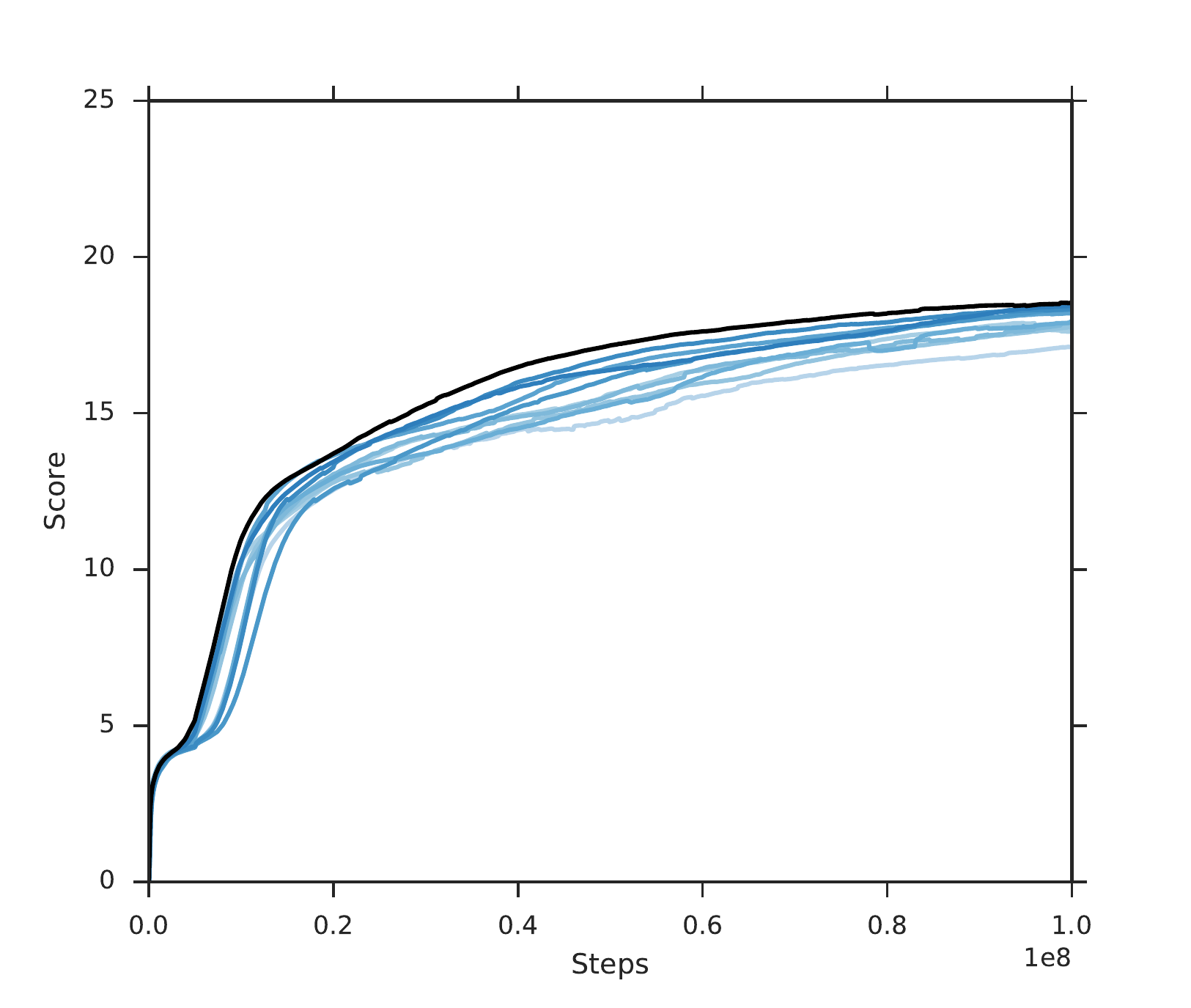}
  \end{subfigure}
  \begin{subfigure}[h]{0.405\linewidth}
    \includegraphics[width=\linewidth]{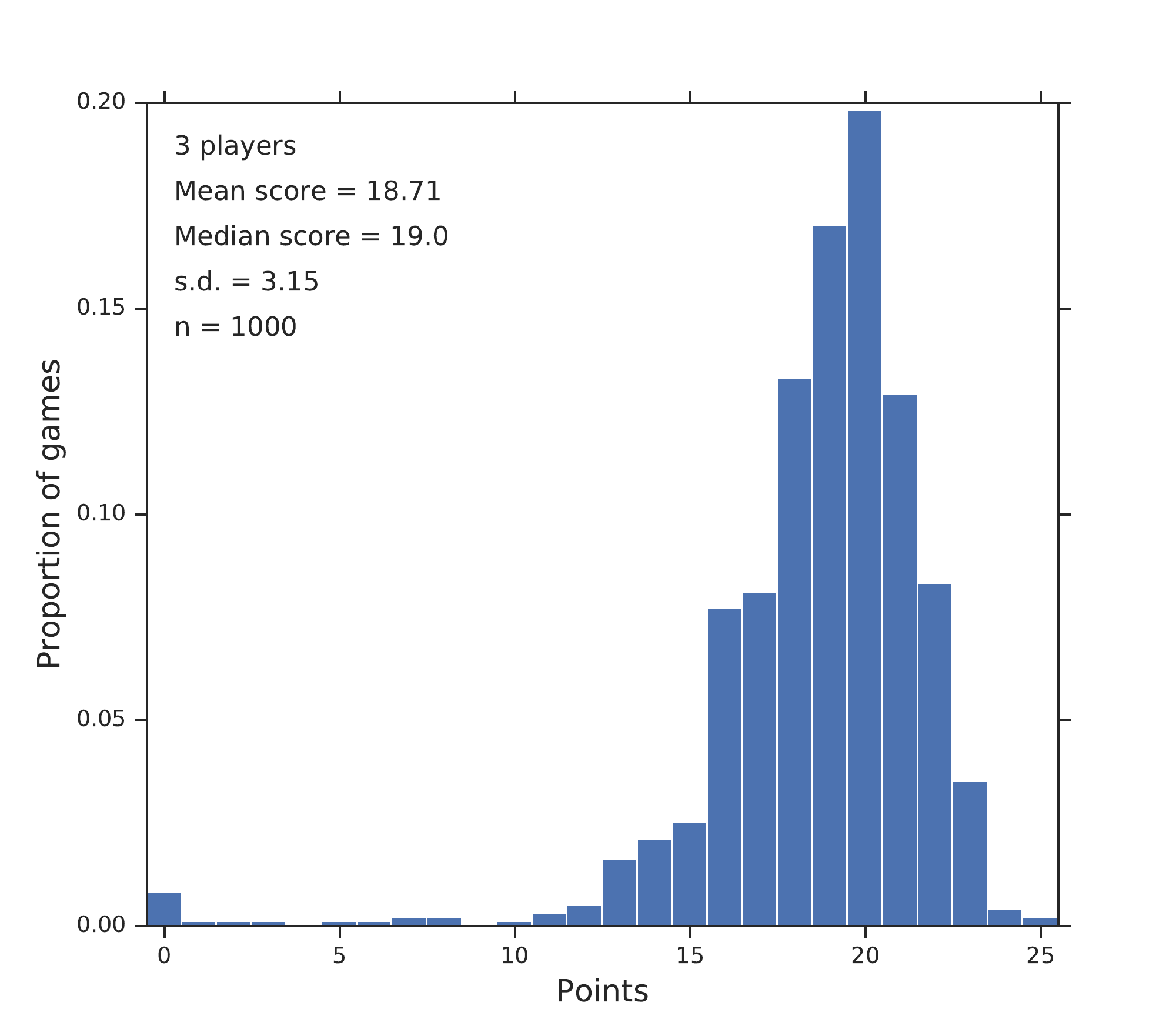}
  \end{subfigure}
  \vskip -4pt
  \begin{subfigure}[h]{0.405\linewidth}
    \includegraphics[width=\linewidth]{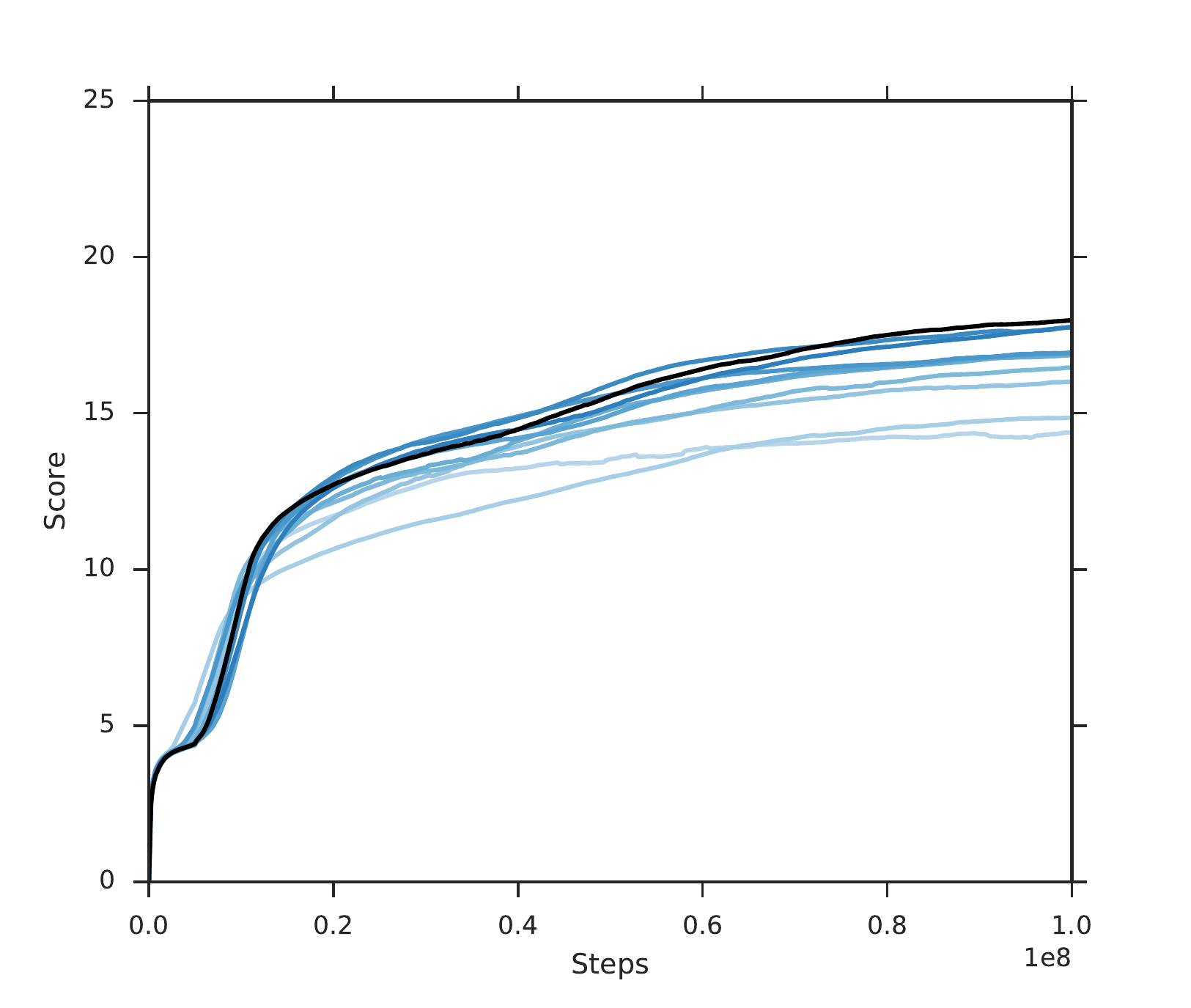}
  \end{subfigure}
  \begin{subfigure}[h]{0.405\linewidth}
    \includegraphics[width=\linewidth]{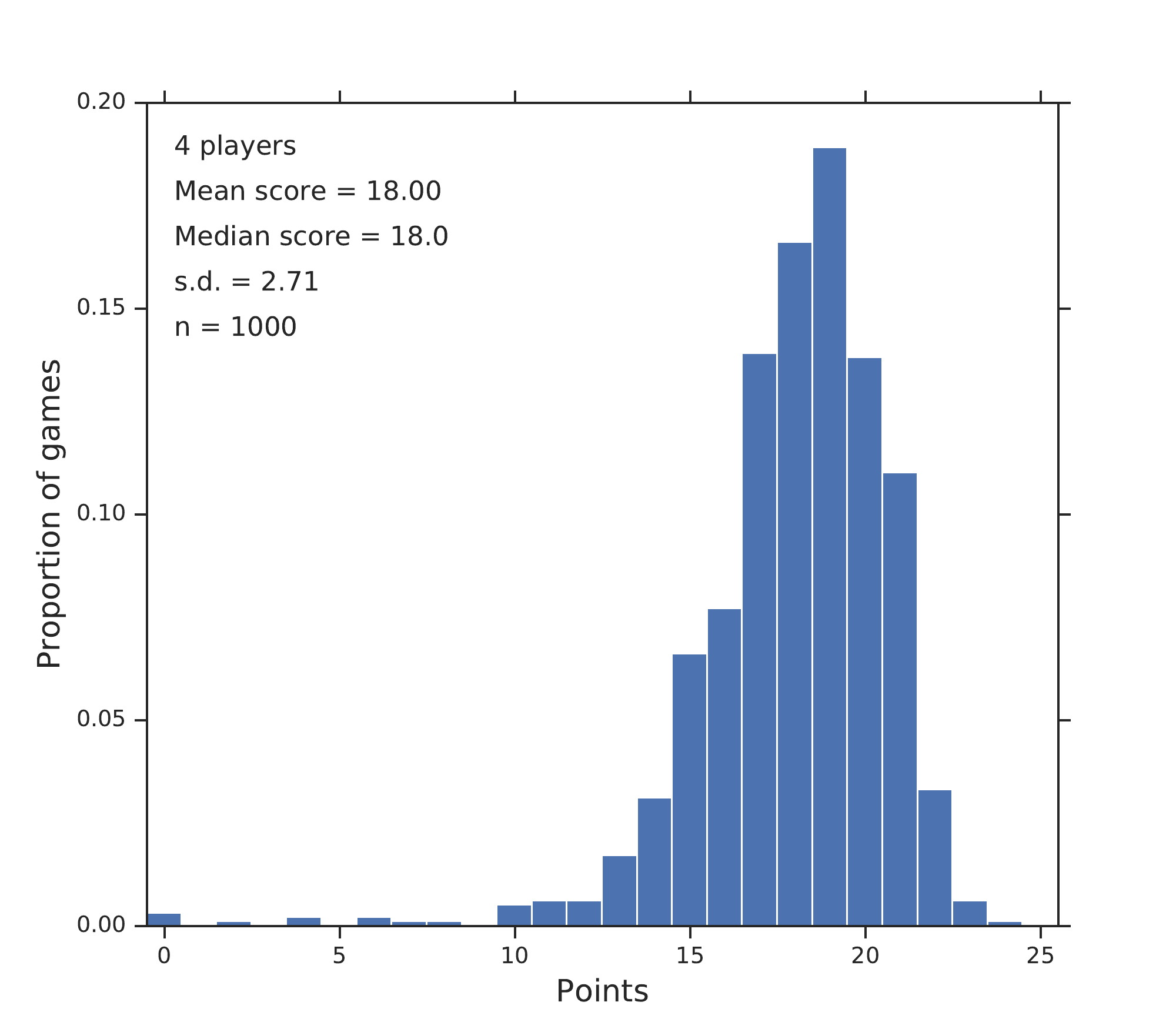}
  \end{subfigure}
  \vskip -4pt
  \begin{subfigure}[h]{0.405\linewidth}
    \includegraphics[width=\linewidth]{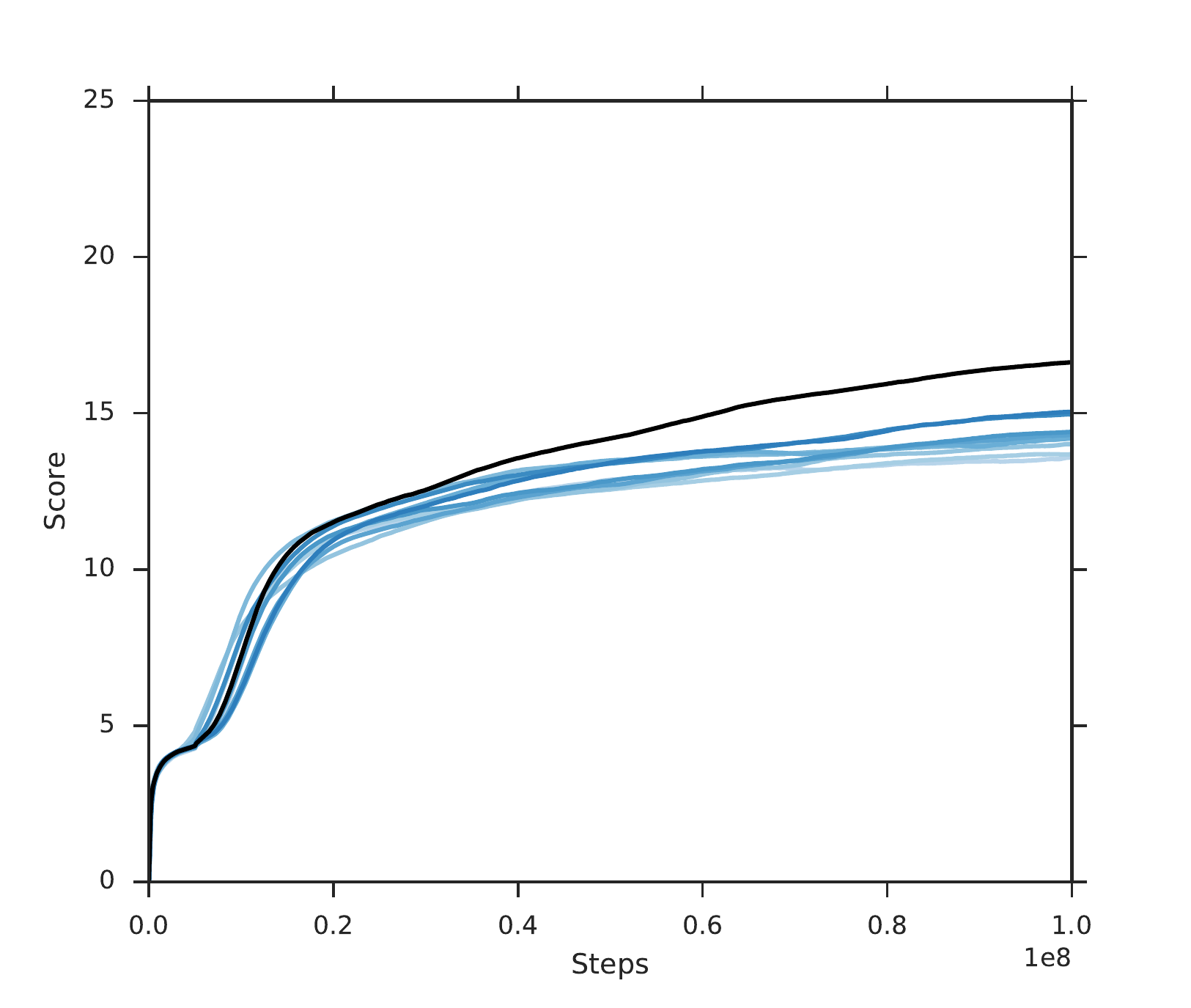}
  \end{subfigure}
  \begin{subfigure}[h]{0.405\linewidth}
    \includegraphics[width=\linewidth]{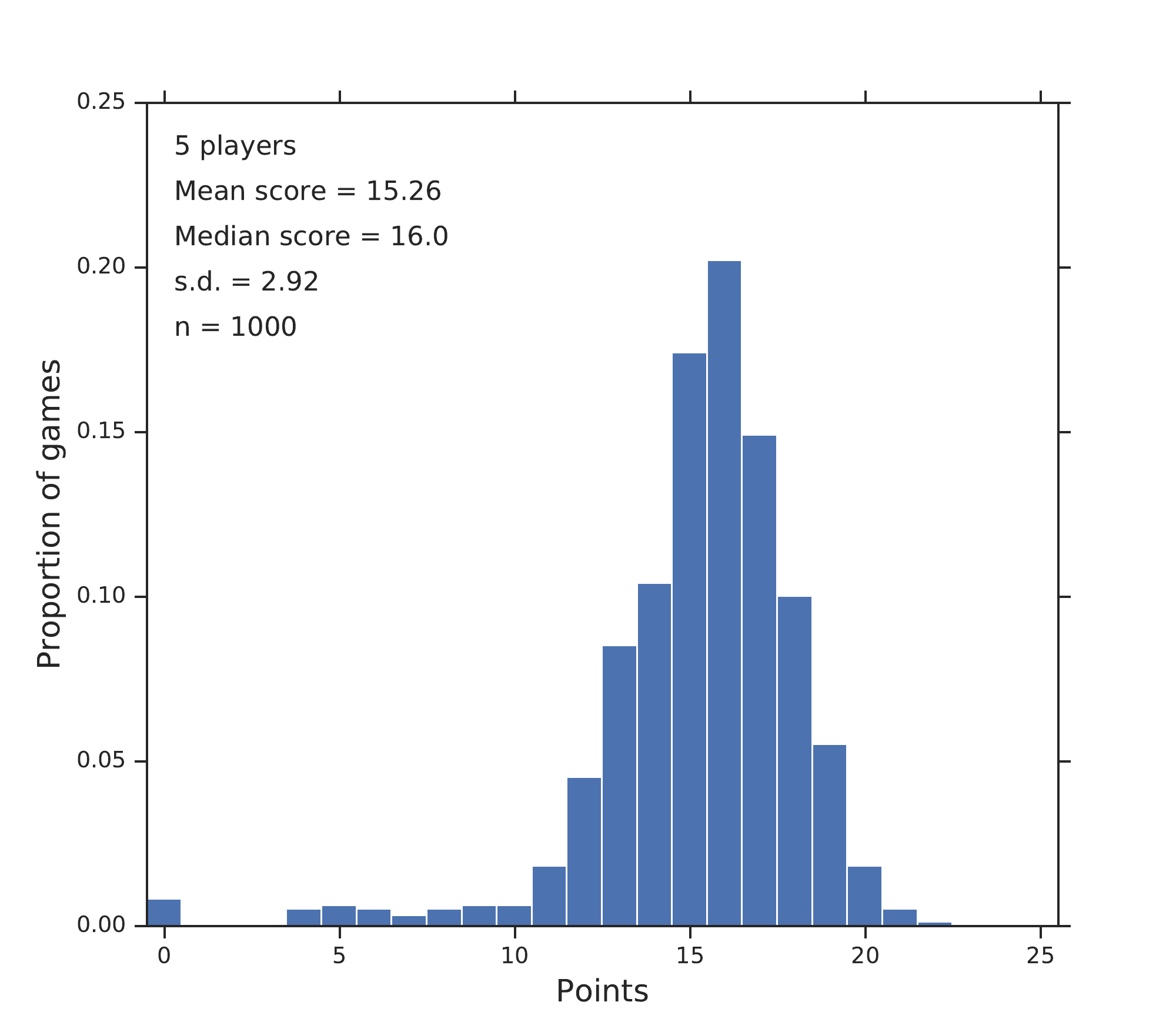}
  \end{subfigure}
  \caption{Rainbow results for two to five players, from top to bottom
        respectively. Performance curves (left) are training-time results from
        the current policy. Average scores and distributions (right) are
        test-time results from 1000 episodes generated using the agent with the
        best training score. $\epsilon$ in $\epsilon$-greedy for all agents is
        set to zero.}
  \label{fig:rainbow-results}
\end{figure}

In contrast to \acha{}, the Rainbow agents exhibit low variance across
different independent training runs, as shown by the learning curves
in Figure~\ref{fig:rainbow-results}.   In this case each line
represents an independent training trial.
In addition, Rainbow agents tend to converge to similar strategies,
seemingly identifying the same local minima. 
In particular, Rainbow agents are 3-4 times less likely to hint for color than \acha{}, and when they do there is no evidence of specific conventions associated with the color hinted. Instead all Rainbow agents we looked at primarily hint for rank, typically indicating that the most recent card is playable.
See \ref{app:conditionalprob} for details.
We speculate that two factors contribute to this consistency across
different runs.  First, Rainbow has a one-step memory for the past action and no memory of past observations. This drastically reduces the space of possible strategies.
Second, Rainbow is a value-based method and starts out with a high
exploration rate.  During the initial exploration, since agents fail
to successfully finish any games without running out of lives,
Q-values will tend towards zero. This might cause agents to learn
from effectively the same starting Q-values, even across independent
runs.

\begin{figure}[h]
  \centering
  \begin{subfigure}[h]{0.48\linewidth}
    \includegraphics[width=\linewidth]{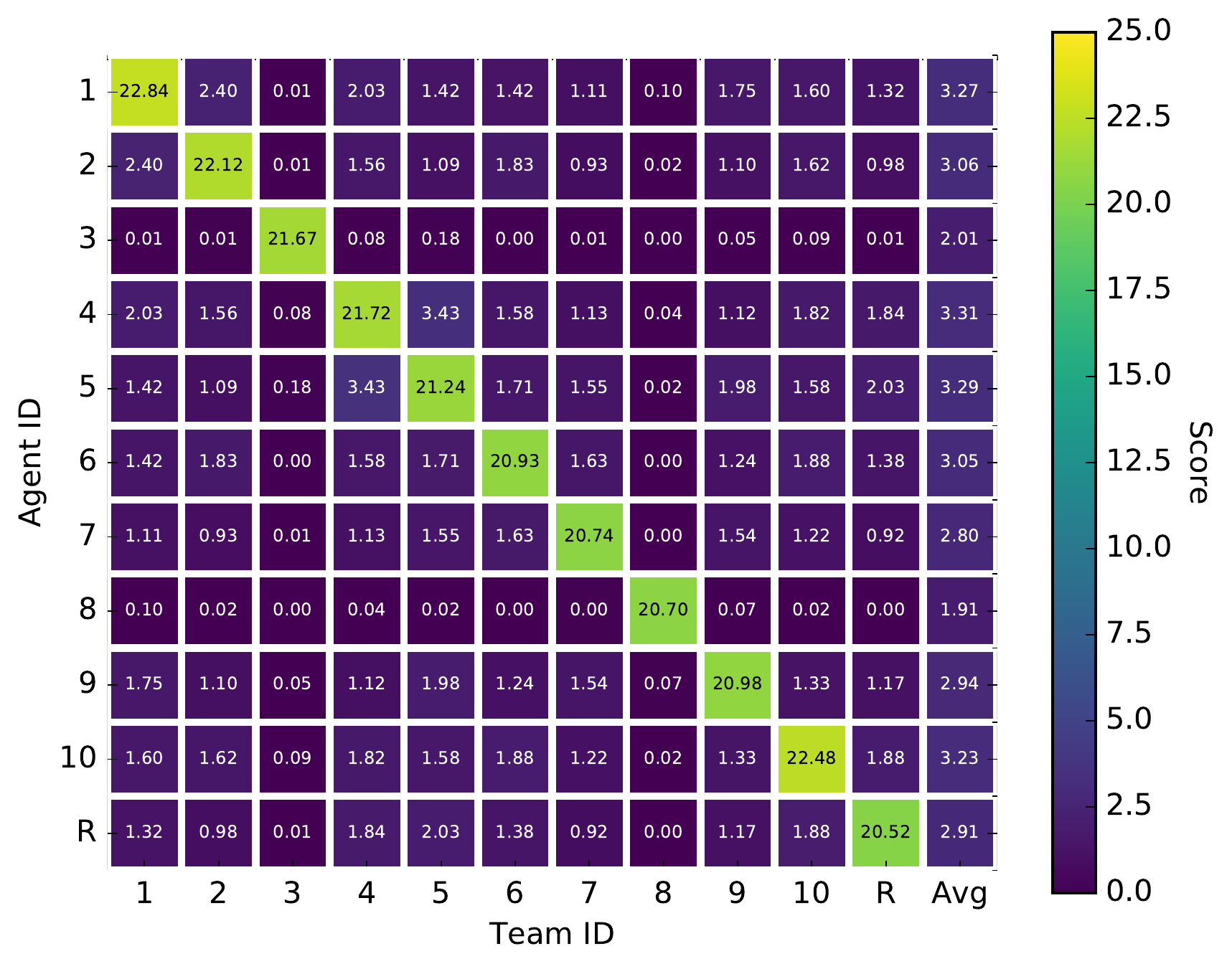}
      \captionsetup{font=scriptsize}
      \caption{Ad-hoc results for two players.}
      \label{fig:firebeast-results-mixing-2p-noEvolve-strategies}
  \end{subfigure}
  \begin{subfigure}[h]{0.48\linewidth}
    \includegraphics[width=\linewidth]{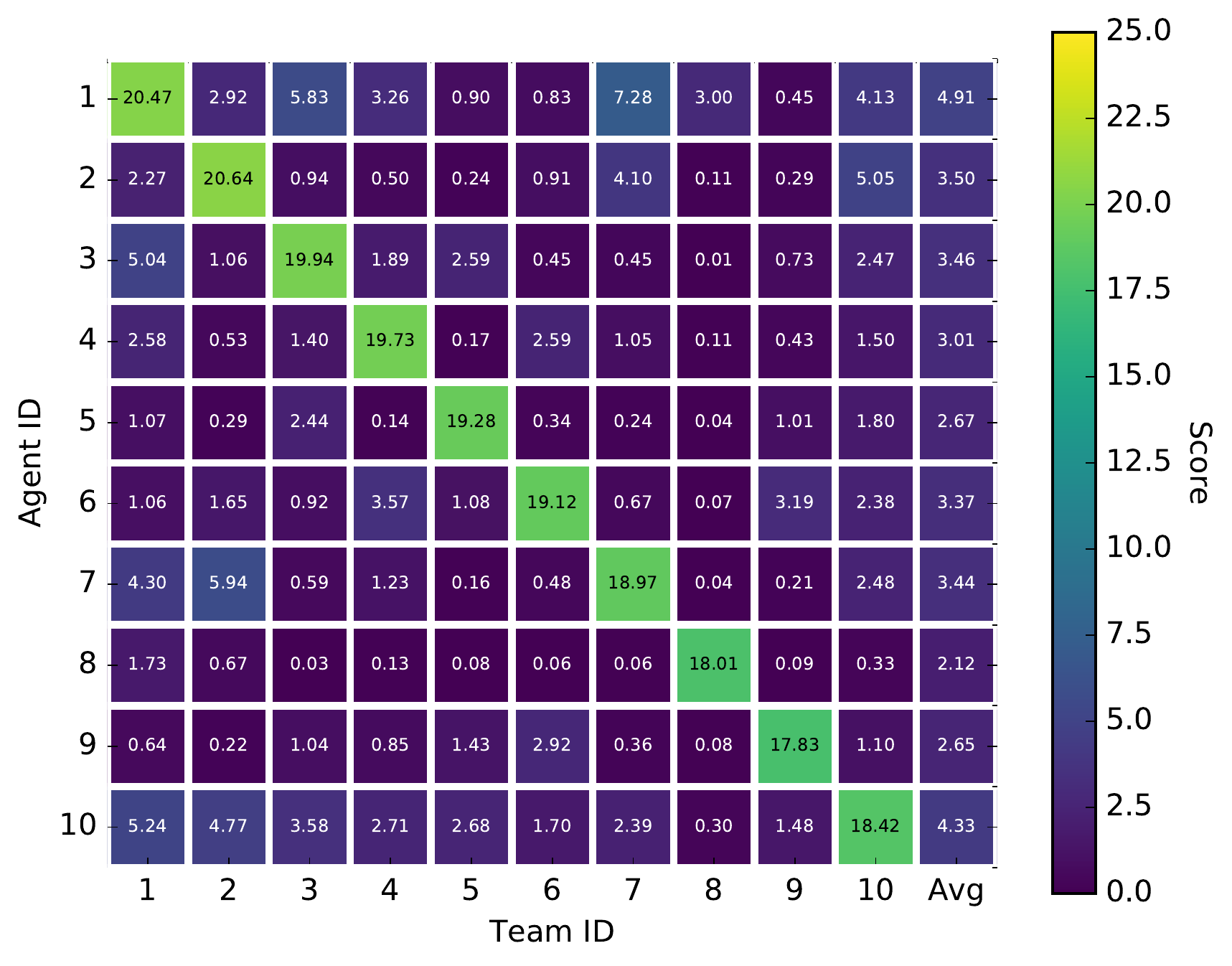}
      \captionsetup{font=scriptsize}
      \caption{Ad-hoc results for four players.}
      \label{fig:firebeast-results-mixing-4p-noEvolve-strategies}
  \end{subfigure}
        \caption{Ad-hoc team results. Teams were constructed using the 10 best independently trained \acha{} agents (\textbf{1--10}; see Figure~\ref{fig:firebeast-results-no-evolution}) and the best Rainbow agent (\textbf{R}; see Figure~\ref{fig:rainbow-results}). Mean scores over 1000 trials are shown. Standard error of the mean was less than $0.09$ and $0.255$ for two and four players, respectively.}
  \label{fig:firebeast-results-mixing-selfplay-strategies}
\end{figure}

\subsection{Experimental Results: Ad-Hoc Team Play}
\label{sec:experiments:ad-hoc}


The policies learned by the aforementioned \acha{}, BAD  and Rainbow agents
are moderately effective in self-play. We now investigate these
agents in the ad-hoc team setting, where teammates are using different
conventions. In particular, we examine the
performance of different ad-hoc teams of \acha{}  and Rainbow
agents. 
Since we established in
Section~\ref{sec:experiments:self-play} that Rainbow agents learn
nearly identical strategies across different runs, for the rest of the
section we consider a population of agents made up of the best
performing Rainbow agent and a  pool of independently trained \acha{}
agents taken from the top ten agents according to final training-time performance from Figure~\ref{fig:firebeast-results-no-evolution}.

We do not use the hand-coded agents in the ad-hoc setting, for a mix
of practical and technical reasons. As written, none of the hand-coded
agents interface with the learning environment or are even written in
the same language, with a different non-trivial (and possibly
error-prone) translation required to have the learning agents play
with each hand-coded agent. We made an early effort to play games with
combinations of ACHA agents and SmartBot, but these games always ended
with SmartBot crashing because ACHA makes moves SmartBot assumes
should never be made. Simple fixes only led to a score of zero
points. Given the effort to connect different agents, and the
complicated action coding of agents like HatBot and WTFWThat, we
assume without further evidence that other combinations of hand-coded
agents would have equally bad performance.

As neither of \acha{} and Rainbow make use of sample play of
their ad-hoc teammates, the ad-hoc team's performance will simply
depend on the compatibility of the different protocols.  The purpose
of this section, then, is primarily to illustrate the difficulty in
ad-hoc team play, and suggest a source for creating a diverse pool of
agents for future evaluation.

\noindent\textbf{Two Players.}
Figure~\ref{fig:firebeast-results-mixing-2p-noEvolve-strategies}
shows a crosstable of the agents' test performance in the ad-hoc setting.
The entry for the $i$-th row and $j$-th column shows the mean
performance of agent $i$ playing in an ad-hoc team consisting of agent $j$,
evaluated for 1000 games with
a random player starting each game. The scores of 20 or greater along
the diagonal entries show that the agents indeed perform well in
self-play. However, when paired with other agents, performance drops
off sharply, with some agents scoring essentially zero in any of these
ad-hoc teams.

\noindent\textbf{Four Players.}
We observe analogous results when evaluating ad-hoc teams in the four
player setting.  We used the top ten \acha{} agents from
Figure~\ref{fig:firebeast-results-no-evolution}.
Similar to the two player ad-hoc results, the entry for the $i$-th row and
$j$-th column of
Figure~\ref{fig:firebeast-results-mixing-4p-noEvolve-strategies} shows
the mean performance of agent $i$ when playing with an ad-hoc team consisting
of three other agent $j$ players, evaluated for 1000 games with a random player
starting each game.
As in the two player results, the agents fare relatively well in
self-play but performance dramatically decreases once we introduce a
second unique agent to the team.

Developing agents that can learn from, adapt, and play well with unknown teammates represents a formidable challenge.

%% file: core/7_related_work.tex
\section{Hanabi: Related Work}
\label{sec:related}


This challenge connects research from several communities, including
reinforcement learning, game theory, and emergent communication. 
In this section, we discuss prior work on the game of Hanabi.
We also briefly discuss notable work from these other communities.

\subsection{Prior Work on Hanabi AI}
\label{sec:related:hanabi}

To the best of our knowledge, the earliest published work on Hanabi was in 2015.
Osawa~\citep{Osawa15} described some of the unique elements of Hanabi, and showed that simulated strategies that try to
recognize the intention of the other players performed better than a fixed set of static
strategies in two-player Hanabi. Later in the same year, Cox et al.~\citep{Cox15} developed the
hat strategy described in Section~\ref{sec:experiments_rulebased}.

Several studies have focused on techniques to achieve strong Hanabi play. First, van den Bergh
et al. described fixed rules whose trigger thresholds were optimized by manual tuning via
simulation~\citep{vandenBergh16}; the action selection is also improved during play using
Monte Carlo planning. Walton-Rivers and colleagues~\citep{WaltonRivers17} evaluated several rule-based
and Monte Carlo tree search agents and observed a predictor version that
modelled the other
players performed better than bots without this ability.
Finally, Bouzy~\citep{Bouzy17} examined several heuristic players and found that combining
search with the hat strategy yielded the best-performing
agents. In the five-player game when ignoring the rule that
hints cannot refer to an empty set of cards, they reported achieving 24.92
points and a perfect score 92\% of the time on average.  While some of these
ideas are used in the agents that we describe in
Section~\ref{sec:experiments_rulebased}, the agents we benchmark have
superior performance under the complete Hanabi rules to the numbers reported in these papers. 



There are three related works that are not about (directly) increasing
performance.  Baffier and colleagues~\citep{Baffier17} examine the complexity of the generalized
game and found optimal gameplay in Hanabi to be NP-hard, even with a
centralised cheating player playing all seats with perfect information of all
cards, including the order of cards in the deck.  However,
without a centralized cheating player, Hanabi is an instance of a decentralized
Markov decision process (DEC-MDP) since the players jointly observe the full
state of the game. Bernstein and colleagues~\citep{BernsteinGIZ02:DEC-MDP}
showed that solving DEC-MDPs is in the nondeterministic exponential time (NEXP)
complexity class, i.e., requiring exponential time even if P=NP.
Liang and colleagues~\citep{LiangPAK19} compared the performance
of artificial agents that implicate additional meaning through their hints with
agents that simply hint to minimize entropy, and found that humans were more
likely to believe the implicature-based agent was also a
human.
Eger and Martens~\citep{Eger17} describe how to model knowledge and how it is revealed through actions using dynamic
epistemic logic. These epistemic formalizations within game theory attempt to
quantify what players know and assume others know, how players can reason about this knowledge,
and what rationality means in this context~\citep{Pacuit17}. These ideas are
also reflected in the BAD agent~\cite{Foerster18BAD}, where they are
combined with scalable deep multi-agent RL, and have resulted in the
strongest two-player Hanabi agent.


Recently, there was a Hanabi competition that took place at the 2018 IEEE Computational
Intelligence in Games Conference in Maastricht~\citep{CIG2018Competition}. There were a total of five agents, three
submissions and two samples . There were two tracks, called
``Mirror'' and ``Mixed'' which broadly match the two categories we propose in
Section~\ref{sec:benchmark}.
Similarly to van den Bergh, the second-place player used a genetic algorithm to evolve a 
sequence of rules from a fixed rule set~\citep{Canaan18}. This agent achieved an average
score of 17.52 points in the Mirror competition, while the first place agent, ``Monte Carlo NN'',  achieved a
score of 20.57. According to the website these scores are averaged
across the results for two to five players.
The winning agent, developed by James Goodman~\citep{Goodman19}, won both the Mirror and Mixed tracks; it used Information Set Monte Carlo
Tree Search~\citep{Cowling12ISMCTS} as
a base algorithm coupled with a re-determinization technique that re-samples consistent world states for
everyone but the player acting at a node. This entry also used a neural network to represent a policy and
value function trained via self-play.
Goodman contrasts the self-play performance of this approach using
different configurations with our results and prior works under each of the two
through five player settings.  Like many prior Hanabi agents, Goodman's
approach exploited some fixed rules and conventions: selecting actions from a
fixed set of nine rules and using the convention that cards are playable when
hinted.  In the Mixed track, the rollouts of other agents incorporated some
agent modelling by using Bayesian predictions about which of a set of policies
the other agents were using.  Although Goodman's approach was fairly effective
in the Mixed track, it is unclear how well it would fare with other agents,
particularly those that are not built using published rules or rules encoded by
the competition's code framework.  Critically, like many prior works on
Hanabi, the competition scored failed games (where all the lives were used) as
the number of cards successfully played prior to failure, whereas we give such
failed games a score of zero.  While this distinction may have relatively
little impact in self-play, as there is no risk of misinterpreting actions, our
experiments suggested this has considerable impact in the ad-hoc teamwork
setting.

\subsection{Reinforcement Learning (RL) and Multi-Agent RL}


While there has been much work in rule-based and search-based players using various heuristics based
on domain knowledge, we are unaware of any prior approaches on learning to play Hanabi directly
from experience given only the rules of the game.
The framework for this approach is {\it reinforcement learning}, where an agent chooses actions in its
environment, receives observations and rewards~\citep{SuttonBarto18}. The goal is to learn a
policy that achieves a high expected sum of rewards, \emph{i.e.} score.

The setting with multiple reinforcement learning agents was first investigated in competitive
games \citep{Littman94markovgames}. This was the start of the foundational work on {\it multiagent}
reinforcement learning (MARL), which focused on algorithms and convergence properties
in Markov/stochastic (simultaneous move) games.
Independent learners in the cooperative case, even in these simpler game models, already 
face several coordination problems~\citep{Matignon12}.
Several years of work on MARL gave rise to many algorithms,
extensions, and analyses~\citep{bowling2003simultaneous,Panait05,Busoniu08Comprehensive,Nowe12Game,BloembergenTHK15}.
For more on multi-agent deep reinforcement learning, Hernandez-Leal and
colleagues~\citep{HernandezLeal18Survey} provide a recent survey.

Model-free methods, which learn a direct mapping from observations to actions, are
appealing in traditional RL because the agent need not understand the dynamics of the environment
in order to act. In games, however, the perfect model (\emph{i.e.}, the rules) is given. This leads to
methods than can combine planning with RL, as first demonstrated by TD-Leaf~\citep{Baxter98} and
TreeStrap~\citep{VenessSilver09}. Recently, Monte Carlo tree search~\citep{kocsis06bandit} was
combined with deep neural networks in AlphaGo~\citep{Silver16Go}, a computer Go-playing agent that was stronger
than any preceding program and defeated the best human Go players.
Shortly thereafter AlphaGo was surpassed by AlphaGo Zero, which used less knowledge~\citep{Silver17AGZ},
and then AlphaZero which also learned to play chess and shogi at super-human
levels~\citep{Silver18AZ}.

Many of the above successes were in applications of RL to perfect information
games (\emph{e.g.}, Go, chess, shogi).
Imperfect-information environments offer new challenges since the world cannot be perfectly simulated:
agents must reason about information that is not known to them. In the competitive case, such as
poker, strong play required new algorithmic
foundations~\citep{CFR,Bowling15Poker,Brown17Libratus,Moravcik17DeepStack}.

Finally, there has been recent interest in the problem of emergent communication. These problems
include an arbitrary or structured communication channel, and agents must learn to communicate
to solve a cooperative problem.
Algorithms have been developed to learn to solve riddles and referential
games~\citep{Foerster16RIAL,Lazaridou16},
gridworld games requiring coordination~\cite{Sukhbaatar16},
object identification via question-and-answer dialog~\citep{deVries17GuessWhat}, 
and negotiation~\citep{Lewis17,Cao18}.
The main difference in Hanabi is that there is no ``cheap-talk'' communication channel: any signalling must be done
through the actions played. It is therefore more similar in spirit to learning how to relay information
to partners in the bidding round of bridge~\citep{Yeh16}.

\subsection{Pragmatics, Beliefs, and Agent Modelling}

Hanabi provides AI practitioners with an interesting multi-agent learning
challenge where they can explore agent communication not just between
artificial agents, but also with human partners.  Understanding human
communication seems important if artificial agents are to collaborate
effectively with humans, as they will likely need to communicate with us on
our terms.  When communicating with natural language, humans exploit theory of
mind through their use of \emph{pragmatics}: conveying meaning based on not
only what was literally said, but also what is implicated (\emph{i.e.},
suggested) by the speaker based on the context~\cite{grice1975:logic}.  By
relying on a listener to disambiguate a speaker's intended meaning through
mechanisms like pragmatic reasoning, communication can be more
efficient~\cite{PIANTADOSI2012280}, as in the case with humans 
hinting about playable cards in Hanabi.  Notably, computational models of human
pragmatic reasoning depend on speakers being cooperative and having the
\emph{intention} of communicating useful information~\cite{Frank998}.

There has been much work that proposes to model beliefs about the intentions or plans of other agents. Several formalisms have been proposed for this, such as I-POMDPs~\cite{Gmytrasiewicz05}
and Bayesian games~\citep{Shoham09}. However, algorithms to compute exact solutions quickly become
intractable for complex problems.

Classical models to predict human behaviour in games include ways to deal with imperfect
rationality~\citep{Yoshida08,Wright17}.
In recent years,
several mechanisms have been proposed to learn these models using deep learning, from human
data in one-shot games~\citep{Hartford16}, learning players with expert features~\citep{DRON},
and end-to-end prediction of fixed policies in gridworld games~\citep{Rabinowitz18}.
When predictions are used to exploit or coordinate, this is often called
{\it opponent/teammate modelling} or more generally {\it agent modelling}~\citep{Albrecht18Modeling}.
Several MARL algorithms have been recently proposed that learn from models of
others whether by using
a training regime based on cognitive hierarchies~\citep{Lanctot17PSRO}, by defining a
learning rule that shapes the anticipated learning of other agents~\citep{Foerster17LOLA},
or by training architectures that incorporate agent identification as part of the objective and
conditioning the policy on these predictions~\citep{Grover18}.
Lastly, an approach in poker (DeepStack) introduced function approximators that represent
belief-conditional values and explicitly representing and reasoning
about manipulating agents' beliefs ~\citep{Moravcik17DeepStack},
similar to the approach of BAD~\cite{Foerster18BAD}.

%% file: core/8_conclusion.tex
\section{Conclusion}

The combination of cooperative gameplay and imperfect information make Hanabi a
compelling research challenge for machine learning techniques in multi-agent
settings.  We evaluate state-of-the-art reinforcement learning algorithms using
deep neural networks and demonstrate that they are largely insufficient to
even surpass current hand-coded bots when evaluated in a self-play setting.
Furthermore, we show that in the ad-hoc team setting, where agents must play
with unknown teammates, such techniques fail to collaborate at all.
Theory of mind appears to play an important role in how humans learn
and play Hanabi.  We believe improvements to both learning in self-play and
adapting to unknown teammates will help us understand better the role
theory of mind reasoning might play for AI systems that learn to
collaborate with other agents and humans.
To promote effective and consistent comparison between techniques, we
provide a
new open source code framework for Hanabi and propose evaluation methodology for
practitioners.


%% file: core/acks.tex
\section*{Acknowledgements}
We would like to thank many people:
Matthieu d'Epenoux of Cocktail Games and Antoine Bauza, who designed Hanabi, for their support on this project;
Alden Christianson for help with coordinating across three different time zones, and discussions with Cocktail Games;  
Angeliki Lazaridou for discussion and feedback on pragmatics and theory of mind;
Ed Lockhart for many discussions on small Hanabi-like games and belief-based reasoning;
Daniel Toyama for assistance with writing clear, readable code for the Hanabi research environment used in our experiments;
Kevin Waugh for helpful comments and feedback;
David Wu for providing details on the FireFlower bot;
and Danny Tarlow for early feedback on the project.

%% file: core/appendix.tex
\subsection{FireFlower Details}
\label{app:fireflower}

Implemented conventions include the following:

\begin{itemize}
\item Hints generally indicate play cards, and generally newer cards first.
\item Hints ``chain'' on to other hints, e.g., if $A$ hints to $B$ a playable \textbf{red 2} as \textbf{red}, then $B$ might infer that it is indeed the \textbf{red 2}, and then hint back to $A$ a \textbf{red 3} in $A$'s hand as red, expecting $A$ to believe it is a \textbf{red 3} to play it \emph{after} $B$ plays its \textbf{red 2}.
\item When discarding, discard provably useless cards, otherwise the oldest ``unprotected'' card.
\item Hints about the oldest unprotected card ``protect'' that card, with many exceptions where it instead means play.
\item Hints about garbage cards indicate ``protect'' cards older than it.
\item Deliberately discarding known playable cards signals to the partner that they have a copy of that card (with enough convention-based specificity on location that they may play it with absolutely no actual hints).
\item In many cases, hints about cards that have already been hinted change the
belief about those cards in various ways such that the partner will likely
change what they do (in accordance with the broader heuristic that one usually
should give a hint if and only if they would like to change the partner's
future behaviour).
\end{itemize}

\subsection{Conditional Probability Tables}
\label{app:conditionalprob}

The tables below show conditional probability summaries of the learned policies from different runs of ACHA and Rainbow in the two player game.  See Section~\ref{sec:experiments} for a discussion of these policy summaries.

\begin{sidewaystable}[h]
  {\scriptsize
  \centering
  \begin{tabular}{|ll|rrrrr|rrrrr|rrrrr|rrrrr|}
    \hline
\multicolumn{2}{|l|}{$P_\%(a_{t+1}|a_t)$} & \multicolumn{5}{l|}{Discard} & \multicolumn{5}{l|}{Play} & \multicolumn{5}{l|}{Hint Colour} & \multicolumn{5}{l|}{Hint Rank} \\
& & 1 & 2 & 3 & 4 & 5 & 1 & 2 & 3 & 4 & 5 & R & Y & G & W & B & 1 & 2 & 3 & 4 & 5 \\
\hline
\multicolumn{2}{|l|}{$a_t$} & \multicolumn{20}{c|}{$a_{t+1}$} \\
\hline
Discard & 1 & 4 & 4 & 3 & 1 & 10 & 2 & 1 & 1 & 0 & 1 & 1 & 6 & 2 & 2 & 1 & 12 & 14 & 11 & 12 & 11\\ 
 & 2 & 2 & 6 & 5 & 2 & 12 & 1 & 0 & 0 & 0 & 0 & 1 & 4 & 2 & 2 & 1 & 14 & 18 & 13 & 11 & 5\\ 
 & 3 & 2 & 6 & 11 & 2 & 13 & 1 & 1 & 1 & 1 & 2 & 1 & 3 & 2 & 2 & 1 & 18 & 17 & 11 & 4 & 1\\ 
 & 4 & 1 & 2 & 5 & 4 & 19 & 0 & 1 & 0 & 0 & 0 & 1 & 6 & 1 & 6 & 2 & 18 & 14 & 10 & 6 & 2\\ 
 & 5 & 1 & 1 & 1 & 0 & 30 & 1 & 0 & 0 & 0 & 0 & 0 & 3 & 2 & 2 & 2 & 10 & 12 & 16 & 13 & 6\\\hline
Play & 1 & 4 & 5 & 3 & 1 & 18 & 4 & 2 & 1 & 1 & 2 & 0 & 2 & 1 & 1 & 1 & 13 & 10 & 10 & 11 & 9\\ 
 & 2 & 4 & 6 & 5 & 1 & 22 & 3 & 2 & 1 & 0 & 1 & 0 & 2 & 1 & 1 & 0 & 11 & 12 & 11 & 11 & 6\\ 
 & 3 & 4 & 5 & 2 & 1 & 26 & 4 & 2 & 2 & 1 & 2 & 0 & 3 & 1 & 1 & 1 & 7 & 6 & 8 & 14 & 10\\ 
 & 4 & 3 & 3 & 4 & 0 & 0 & 3 & 5 & 1 & 1 & 1 & 0 & 3 & 0 & 0 & 0 & 49 & 10 & 1 & 12 & 3\\ 
 & 5 & 2 & 4 & 5 & 1 & 26 & 2 & 1 & 0 & 0 & 0 & 0 & 3 & 1 & 1 & 0 & 12 & 14 & 12 & 10 & 5\\\hline
Hint & R & 6 & 31 & 25 & 9 & 3 & 0 & 0 & 0 & 0 & 1 & 9 & 4 & 2 & 5 & 2 & 1 & 2 & 0 & 1 & 1\\ 
Colour & Y & 19 & 31 & 12 & 5 & 8 & 1 & 0 & 1 & 0 & 0 & 1 & 7 & 2 & 3 & 1 & 2 & 3 & 2 & 2 & 2\\ 
 & G & 20 & 30 & 18 & 1 & 9 & 2 & 0 & 1 & 0 & 0 & 0 & 4 & 3 & 1 & 1 & 0 & 2 & 2 & 2 & 2\\ 
 & W & 18 & 18 & 16 & 10 & 15 & 0 & 0 & 0 & 0 & 0 & 1 & 4 & 1 & 8 & 0 & 2 & 2 & 1 & 2 & 1\\ 
 & B & 12 & 8 & 11 & 6 & 37 & 2 & 0 & 0 & 0 & 0 & 0 & 2 & 2 & 4 & 3 & 1 & 1 & 3 & 3 & 4\\\hline
Hint & 1 & 3 & 3 & 6 & 8 & 1 & 8 & 9 & 0 & 8 & 50 & 0 & 0 & 0 & 0 & 0 & 3 & 0 & 1 & 0 & 0\\ 
Rank & 2 & 3 & 6 & 1 & 0 & 0 & 8 & 23 & 0 & 0 & 52 & 0 & 0 & 0 & 0 & 0 & 3 & 1 & 0 & 0 & 0\\ 
 & 3 & 1 & 5 & 2 & 0 & 1 & 11 & 20 & 10 & 0 & 47 & 0 & 0 & 0 & 0 & 0 & 0 & 1 & 1 & 0 & 0\\ 
 & 4 & 1 & 4 & 3 & 1 & 1 & 13 & 22 & 6 & 1 & 37 & 0 & 0 & 0 & 0 & 0 & 6 & 2 & 1 & 1 & 0\\ 
 & 5 & 2 & 4 & 0 & 0 & 2 & 29 & 19 & 7 & 2 & 28 & 0 & 1 & 0 & 0 & 0 & 0 & 1 & 1 & 1 & 2\\\hline \hline
\multicolumn{2}{|l|}{$P_\%(a_t)$}  &  3 & 5 & 4 & 2 & 14 & 5 & 7 & 2 & 1 & 17 & 0 & 2 & 1 & 1 & 1 & 9 & 8 & 7 & 7 & 4\\
\hline 
  \end{tabular}
  \caption{Conditional action probabilities for a first two player Rainbow agent.}
  }
\end{sidewaystable}

\begin{sidewaystable}[h]
  \centering
  \scriptsize
  \begin{tabular}{|ll|rrrrr|rrrrr|rrrrr|rrrrr|}
    \hline
\multicolumn{2}{|l|}{$P_\%(a_{t+1}|a_t)$} & \multicolumn{5}{l|}{Discard} & \multicolumn{5}{l|}{Play} & \multicolumn{5}{l|}{Hint Colour} & \multicolumn{5}{l|}{Hint Rank} \\
& & 1 & 2 & 3 & 4 & 5 & 1 & 2 & 3 & 4 & 5 & R & Y & G & W & B & 1 & 2 & 3 & 4 & 5 \\
\hline
\multicolumn{2}{|l|}{$a_t$} & \multicolumn{20}{c|}{$a_{t+1}$} \\
\hline
Discard & 1 & 4 & 1 & 1 & 1 & 22 & 1 & 1 & 1 & 0 & 1 & 1 & 2 & 3 & 2 & 0 & 10 & 21 & 9 & 10 & 7\\ 
 & 2 & 2 & 3 & 1 & 0 & 19 & 1 & 1 & 1 & 1 & 1 & 2 & 2 & 3 & 3 & 0 & 9 & 14 & 19 & 11 & 9\\ 
 & 3 & 4 & 1 & 6 & 1 & 19 & 2 & 0 & 0 & 0 & 2 & 3 & 3 & 4 & 3 & 3 & 21 & 11 & 9 & 2 & 5\\ 
 & 4 & 3 & 1 & 1 & 2 & 29 & 1 & 0 & 1 & 0 & 0 & 2 & 3 & 3 & 6 & 2 & 12 & 15 & 11 & 5 & 4\\ 
 & 5 & 1 & 1 & 0 & 0 & 32 & 1 & 0 & 0 & 0 & 0 & 0 & 3 & 2 & 4 & 0 & 10 & 13 & 14 & 11 & 5\\\hline
Play & 1 & 4 & 2 & 1 & 1 & 24 & 3 & 2 & 2 & 0 & 3 & 1 & 1 & 1 & 1 & 0 & 11 & 12 & 10 & 11 & 10\\ 
 & 2 & 4 & 2 & 1 & 1 & 28 & 3 & 2 & 1 & 0 & 2 & 0 & 1 & 1 & 1 & 0 & 12 & 14 & 12 & 8 & 7\\ 
 & 3 & 4 & 2 & 1 & 0 & 23 & 4 & 2 & 1 & 0 & 4 & 1 & 2 & 1 & 2 & 0 & 11 & 13 & 7 & 12 & 9\\ 
 & 4 & 3 & 2 & 0 & 0 & 8 & 4 & 2 & 1 & 0 & 6 & 0 & 0 & 1 & 1 & 0 & 35 & 23 & 5 & 5 & 4\\ 
 & 5 & 3 & 1 & 1 & 1 & 33 & 2 & 2 & 1 & 0 & 1 & 0 & 2 & 2 & 1 & 0 & 11 & 15 & 12 & 8 & 5\\\hline
Hint & R & 15 & 23 & 8 & 9 & 5 & 1 & 0 & 0 & 0 & 0 & 11 & 0 & 7 & 4 & 4 & 3 & 1 & 1 & 1 & 5\\ 
Colour & Y & 28 & 15 & 11 & 9 & 13 & 1 & 1 & 0 & 0 & 1 & 1 & 4 & 3 & 1 & 0 & 2 & 2 & 3 & 2 & 3\\ 
 & G & 22 & 19 & 5 & 4 & 17 & 1 & 1 & 1 & 0 & 1 & 3 & 2 & 8 & 2 & 0 & 2 & 1 & 2 & 2 & 6\\ 
 & W & 9 & 7 & 2 & 4 & 60 & 1 & 1 & 1 & 0 & 1 & 2 & 2 & 3 & 4 & 0 & 1 & 1 & 1 & 1 & 2\\ 
 & B & 17 & 7 & 21 & 17 & 4 & 1 & 0 & 0 & 1 & 5 & 7 & 0 & 0 & 1 & 10 & 1 & 4 & 1 & 2 & 0\\\hline
Hint & 1 & 2 & 3 & 2 & 6 & 1 & 9 & 11 & 6 & 5 & 54 & 0 & 0 & 0 & 0 & 0 & 0 & 0 & 0 & 0 & 0\\ 
Rank & 2 & 8 & 2 & 1 & 0 & 1 & 7 & 21 & 0 & 0 & 48 & 0 & 0 & 0 & 0 & 0 & 8 & 3 & 0 & 0 & 0\\ 
 & 3 & 2 & 2 & 1 & 0 & 1 & 12 & 18 & 11 & 0 & 48 & 0 & 0 & 0 & 0 & 0 & 2 & 1 & 1 & 0 & 0\\ 
 & 4 & 1 & 1 & 0 & 0 & 1 & 24 & 14 & 9 & 1 & 43 & 0 & 0 & 0 & 0 & 0 & 0 & 0 & 1 & 1 & 1\\ 
 & 5 & 8 & 4 & 4 & 2 & 2 & 15 & 19 & 7 & 2 & 23 & 0 & 1 & 2 & 0 & 0 & 2 & 1 & 1 & 3 & 3\\\hline \hline
\multicolumn{2}{|l|}{$P_\%(a_t)$}  &  4 & 2 & 1 & 1 & 19 & 5 & 7 & 3 & 1 & 17 & 0 & 1 & 2 & 2 & 0 & 8 & 9 & 8 & 6 & 4\\
\hline 
  \end{tabular}
  \caption{Conditional action probabilities for second two player Rainbow agent.}
\end{sidewaystable}

\begin{sidewaystable}[h]
  \centering
  \scriptsize
  \begin{tabular}{|ll|rrrrr|rrrrr|rrrrr|rrrrr|}
    \hline
\multicolumn{2}{|l|}{$P_\%(a_{t+1}|a_t)$} & \multicolumn{5}{l|}{Discard} & \multicolumn{5}{l|}{Play} & \multicolumn{5}{l|}{Hint Colour} & \multicolumn{5}{l|}{Hint Rank} \\
& & 1 & 2 & 3 & 4 & 5 & 1 & 2 & 3 & 4 & 5 & R & Y & G & W & B & 1 & 2 & 3 & 4 & 5 \\
\hline
\multicolumn{2}{|l|}{$a_t$} & \multicolumn{20}{c|}{$a_{t+1}$} \\
\hline
Discard & 1 & 9 & 1 & 2 & 1 & 15 & 1 & 1 & 0 & 0 & 0 & 3 & 2 & 2 & 0 & 0 & 19 & 19 & 11 & 9 & 5\\ 
 & 2 & 8 & 4 & 2 & 1 & 16 & 1 & 0 & 0 & 0 & 0 & 6 & 2 & 2 & 1 & 1 & 16 & 20 & 9 & 8 & 2\\ 
 & 3 & 6 & 1 & 2 & 0 & 24 & 1 & 1 & 0 & 0 & 1 & 3 & 3 & 2 & 0 & 1 & 12 & 16 & 14 & 7 & 4\\ 
 & 4 & 7 & 1 & 2 & 2 & 27 & 1 & 1 & 0 & 1 & 1 & 3 & 3 & 2 & 1 & 1 & 14 & 13 & 10 & 7 & 3\\ 
 & 5 & 2 & 0 & 1 & 0 & 32 & 1 & 0 & 0 & 0 & 0 & 3 & 3 & 3 & 0 & 0 & 11 & 12 & 14 & 13 & 4\\\hline
Play & 1 & 9 & 1 & 2 & 1 & 21 & 4 & 3 & 1 & 1 & 2 & 1 & 1 & 0 & 0 & 1 & 10 & 12 & 11 & 12 & 8\\ 
 & 2 & 4 & 1 & 2 & 1 & 25 & 5 & 5 & 1 & 1 & 3 & 1 & 3 & 1 & 0 & 1 & 6 & 7 & 9 & 13 & 12\\ 
 & 3 & 8 & 2 & 2 & 1 & 22 & 2 & 2 & 1 & 1 & 3 & 1 & 1 & 0 & 0 & 0 & 17 & 13 & 10 & 10 & 4\\ 
 & 4 & 5 & 1 & 1 & 2 & 2 & 6 & 3 & 2 & 1 & 8 & 0 & 1 & 0 & 0 & 0 & 33 & 13 & 1 & 15 & 3\\ 
 & 5 & 7 & 1 & 1 & 1 & 29 & 2 & 1 & 1 & 0 & 1 & 2 & 2 & 1 & 0 & 0 & 12 & 15 & 11 & 10 & 4\\\hline
Hint & R & 49 & 15 & 20 & 3 & 4 & 0 & 0 & 0 & 0 & 0 & 0 & 1 & 0 & 0 & 1 & 0 & 3 & 1 & 0 & 1\\ 
Colour & Y & 35 & 5 & 8 & 3 & 12 & 3 & 4 & 1 & 1 & 0 & 1 & 4 & 2 & 0 & 3 & 2 & 2 & 2 & 5 & 7\\ 
 & G & 12 & 4 & 7 & 3 & 55 & 1 & 1 & 0 & 0 & 0 & 1 & 3 & 3 & 1 & 2 & 2 & 1 & 1 & 2 & 1\\ 
 & W & 14 & 13 & 8 & 9 & 27 & 0 & 0 & 0 & 1 & 0 & 2 & 0 & 2 & 12 & 1 & 0 & 2 & 4 & 2 & 1\\ 
 & B & 31 & 2 & 13 & 3 & 12 & 0 & 2 & 1 & 1 & 1 & 2 & 3 & 3 & 1 & 14 & 2 & 1 & 2 & 2 & 3\\\hline
Hint & 1 & 4 & 2 & 5 & 8 & 2 & 10 & 0 & 9 & 6 & 47 & 0 & 0 & 0 & 0 & 0 & 4 & 0 & 1 & 0 & 0\\ 
Rank & 2 & 9 & 0 & 1 & 0 & 0 & 17 & 0 & 15 & 1 & 48 & 0 & 0 & 0 & 0 & 0 & 4 & 2 & 0 & 0 & 0\\ 
 & 3 & 2 & 1 & 1 & 0 & 0 & 14 & 9 & 19 & 0 & 50 & 0 & 0 & 0 & 0 & 0 & 1 & 1 & 1 & 0 & 0\\ 
 & 4 & 3 & 3 & 2 & 1 & 1 & 19 & 14 & 8 & 1 & 38 & 0 & 0 & 0 & 0 & 0 & 5 & 3 & 1 & 2 & 1\\ 
 & 5 & 2 & 1 & 1 & 0 & 2 & 31 & 14 & 8 & 1 & 30 & 1 & 1 & 0 & 0 & 1 & 1 & 0 & 1 & 2 & 3\\\hline \hline
\multicolumn{2}{|l|}{$P_\%(a_t)$}  & 7 & 1 & 2 & 1 & 16 & 7 & 3 & 5 & 1 & 17 & 1 & 2 & 1 & 0 & 0 & 9 & 9 & 7 & 7 & 3\\
\hline 
  \end{tabular}
  \caption{Conditional action probabilities for third two player Rainbow agent.}
\end{sidewaystable}

\begin{sidewaystable}[h]
  \centering
  \scriptsize
  \begin{tabular}{|ll|rrrrr|rrrrr|rrrrr|rrrrr|}
    \hline
\multicolumn{2}{|l|}{$P_\%(a_{t+1}|a_t)$} & \multicolumn{5}{l|}{Discard} & \multicolumn{5}{l|}{Play} & \multicolumn{5}{l|}{Hint Colour} & \multicolumn{5}{l|}{Hint Rank} \\
& & 1 & 2 & 3 & 4 & 5 & 1 & 2 & 3 & 4 & 5 & R & Y & G & W & B & 1 & 2 & 3 & 4 & 5 \\
\hline
\multicolumn{2}{|l|}{$a_t$} & \multicolumn{20}{c|}{$a_{t+1}$} \\
\hline
Discard & 1 & 3 & 4 & 7 & 3 & 6 & 4 & 3 & 1 & 0 & 1 & 6 & 6 & 5 & 6 & 6 & 11 & 7 & 7 & 5 & 9 \\
& 2 & 4 & 5 & 7 & 2 & 6 & 4 & 2 & 2 & 0 & 0 & 6 & 6 & 5 & 6 & 5 & 12 & 6 & 8 & 5 & 9 \\
& 3 & 4 & 4 & 9 & 2 & 7 & 5 & 3 & 1 & 0 & 0 & 5 & 6 & 5 & 6 & 5 & 10 & 6 & 9 & 5 & 8 \\
& 4 & 4 & 4 & 7 & 3 & 12 & 4 & 3 & 1 & 0 & 0 & 4 & 5 & 3 & 4 & 6 & 11 & 8 & 6 & 5 & 9 \\
& 5 & 1 & 2 & 4 & 1 & 9 & 4 & 3 & 1 & 42 & 0 & 2 & 2 & 2 & 2 & 2 & 3 & 6 & 7 & 5 & 2 \\
\hline
Play & 1 & 3 & 3 & 5 & 3 & 6 & 9 & 5 & 2 & 0 & 2 & 4 & 3 & 4 & 5 & 4 & 8 & 8 & 7 & 6 & 11 \\
& 2 & 4 & 4 & 6 & 4 & 7 & 8 & 5 & 1 & 0 & 1 & 4 & 4 & 4 & 5 & 3 & 9 & 7 & 8 & 5 & 11 \\
& 3 & 3 & 3 & 4 & 2 & 7 & 5 & 3 & 1 & 0 & 1 & 5 & 4 & 4 & 4 & 4 & 12 & 15 & 8 & 4 & 9 \\
& 4 & 4 & 4 & 5 & 4 & 9 & 7 & 4 & 1 & 0 & 1 & 4 & 5 & 4 & 5 & 5 & 9 & 9 & 9 & 5 & 5 \\
& 5 & 5 & 5 & 8 & 5 & 10 & 13 & 7 & 1 & 0 & 0 & 3 & 3 & 3 & 3 & 3 & 6 & 7 & 6 & 4 & 5 \\
\hline
Hint & R & 4 & 5 & 3 & 38 & 1 & 16 & 14 & 1 & 0 & 0 & 2 & 1 & 1 & 1 & 2 & 3 & 3 & 2 & 2 & 3 \\
Colour & Y & 4 & 3 & 3 & 45 & 1 & 15 & 12 & 1 & 0 & 0 & 1 & 2 & 1 & 1 & 1 & 3 & 3 & 2 & 1 & 2 \\
& G & 5 & 4 & 2 & 37 & 1 & 19 & 14 & 1 & 0 & 0 & 1 & 1 & 2 & 1 & 2 & 3 & 3 & 2 & 1 & 2 \\
& W & 3 & 3 & 3 & 44 & 1 & 15 & 13 & 1 & 0 & 0 & 2 & 1 & 1 & 1 & 1 & 3 & 2 & 2 & 1 & 2 \\
& B & 3 & 5 & 1 & 44 & 1 & 15 & 12 & 1 & 0 & 0 & 1 & 1 & 1 & 1 & 2 & 3 & 2 & 2 & 2 & 2 \\
\hline
Hint & 1 & 0 & 0 & 0 & 0 & 0 & 9 & 6 & 9 & 12 & 64 & 0 & 0 & 0 & 0 & 0 & 0 & 0 & 0 & 0 & 0 \\
Rank & 2 & 3 & 4 & 2 & 1 & 0 & 2 & 4 & 4 & 22 & 42 & 0 & 0 & 0 & 1 & 0 & 8 & 2 & 1 & 0 & 1 \\
& 3 & 2 & 3 & 1 & 16 & 1 & 5 & 6 & 3 & 26 & 24 & 1 & 1 & 1 & 0 & 1 & 6 & 3 & 1 & 0 & 1 \\
& 4 & 1 & 1 & 0 & 3 & 0 & 2 & 2 & 2 & 55 & 31 & 0 & 0 & 0 & 0 & 0 & 0 & 0 & 0 & 0 & 1 \\
& 5 & 2 & 2 & 2 & 18 & 0 & 6 & 3 & 1 & 55 & 4 & 1 & 0 & 1 & 1 & 1 & 0 & 0 & 1 & 1 & 1 \\
\hline \hline
\multicolumn{2}{|l|}{$P_\%(a_t)$} & 3 & 3 & 4 & 10 & 5 & 8 & 5 & 2 & 10 & 9 & 3 & 3 & 2 & 3 & 3 & 7 & 6 & 5 & 3 & 5 \\
\hline
  \end{tabular}
  \caption{Conditional action probabilities for first two player \acha{} agent.}
\end{sidewaystable}

\begin{sidewaystable}[h]
  \centering
  \scriptsize
  \begin{tabular}{|ll|rrrrr|rrrrr|rrrrr|rrrrr|}
    \hline
\multicolumn{2}{|l|}{$P_\%(a_{t+1}|a_t)$} & \multicolumn{5}{l|}{Discard} & \multicolumn{5}{l|}{Play} & \multicolumn{5}{l|}{Hint Colour} & \multicolumn{5}{l|}{Hint Rank} \\
& & 1 & 2 & 3 & 4 & 5 & 1 & 2 & 3 & 4 & 5 & R & Y & G & W & B & 1 & 2 & 3 & 4 & 5 \\
\hline
\multicolumn{2}{|l|}{$a_t$} & \multicolumn{20}{c|}{$a_{t+1}$} \\
\hline
Discard & 1 & 2 & 2 & 0 & 1 & 31 & 6 & 0 & 0 & 0 & 0 & 9 & 3 & 2 & 3 & 4 & 4 & 2 & 6 & 8 & 16 \\
& 2 & 2 & 2 & 1 & 1 & 33 & 2 & 1 & 1 & 1 & 1 & 8 & 2 & 3 & 4 & 4 & 2 & 2 & 3 & 8 & 17 \\
& 3 & 0 & 2 & 5 & 2 & 0 & 53 & 3 & 4 & 1 & 2 & 5 & 1 & 5 & 2 & 1 & 2 & 2 & 1 & 1 & 10 \\
& 4 & 0 & 0 & 1 & 1 & 0 & 0 & 71 & 0 & 0 & 0 & 4 & 2 & 1 & 1 & 2 & 2 & 2 & 1 & 2 & 11 \\
& 5 & 2 & 1 & 0 & 1 & 31 & 1 & 0 & 0 & 0 & 0 & 11 & 3 & 2 & 6 & 4 & 5 & 3 & 6 & 8 & 16 \\
\hline
Play & 1 & 3 & 3 & 1 & 1 & 27 & 6 & 2 & 1 & 0 & 2 & 7 & 3 & 3 & 3 & 3 & 7 & 2 & 5 & 8 & 12 \\
& 2 & 4 & 3 & 1 & 1 & 27 & 5 & 2 & 1 & 0 & 2 & 6 & 4 & 2 & 4 & 3 & 4 & 3 & 5 & 9 & 13 \\
& 3 & 3 & 2 & 1 & 1 & 25 & 4 & 1 & 1 & 0 & 1 & 8 & 4 & 5 & 4 & 3 & 8 & 3 & 5 & 9 & 12 \\
& 4 & 4 & 2 & 2 & 2 & 31 & 3 & 1 & 0 & 0 & 1 & 7 & 3 & 3 & 7 & 4 & 5 & 3 & 4 & 8 & 11 \\
& 5 & 2 & 1 & 1 & 1 & 33 & 3 & 2 & 0 & 0 & 1 & 11 & 4 & 3 & 5 & 4 & 5 & 3 & 5 & 8 & 9 \\
\hline
Hint & R & 0 & 0 & 0 & 0 & 0 & 0 & 0 & 0 & 0 & 97 & 0 & 0 & 0 & 0 & 0 & 0 & 0 & 0 & 0 & 1 \\
Colour & Y & 0 & 0 & 0 & 1 & 0 & 0 & 94 & 0 & 0 & 0 & 1 & 0 & 0 & 0 & 0 & 1 & 0 & 0 & 0 & 1 \\
& G & 0 & 0 & 0 & 0 & 0 & 0 & 0 & 0 & 99 & 0 & 0 & 0 & 0 & 0 & 0 & 0 & 0 & 0 & 0 & 0 \\
& W & 0 & 0 & 0 & 0 & 0 & 0 & 0 & 0 & 98 & 0 & 0 & 0 & 0 & 0 & 0 & 0 & 0 & 0 & 0 & 0 \\
& B & 0 & 0 & 0 & 1 & 0 & 0 & 95 & 0 & 0 & 0 & 0 & 0 & 0 & 0 & 0 & 1 & 0 & 0 & 0 & 1 \\
\hline
Hint & 1 & 0 & 0 & 0 & 1 & 0 & 23 & 0 & 73 & 0 & 0 & 1 & 0 & 0 & 0 & 0 & 0 & 0 & 0 & 0 & 1 \\
Rank & 2 & 0 & 0 & 0 & 0 & 0 & 47 & 0 & 0 & 0 & 44 & 1 & 1 & 0 & 1 & 1 & 2 & 0 & 1 & 0 & 1 \\
& 3 & 28 & 4 & 1 & 1 & 5 & 43 & 0 & 1 & 0 & 1 & 1 & 1 & 1 & 1 & 1 & 2 & 1 & 2 & 3 & 1 \\
& 4 & 11 & 32 & 1 & 2 & 9 & 17 & 0 & 3 & 0 & 0 & 1 & 2 & 2 & 3 & 2 & 5 & 1 & 3 & 4 & 2 \\
& 5 & 0 & 0 & 0 & 0 & 0 & 0 & 0 & 0 & 0 & 97 & 0 & 0 & 0 & 0 & 0 & 0 & 0 & 0 & 0 & 1 \\
\hline \hline
\multicolumn{2}{|l|}{$P_\%(a_t)$} & 3 & 3 & 1 & 1 & 19 & 6 & 6 & 3 & 5 & 15 & 6 & 2 & 2 & 3 & 3 & 4 & 2 & 4 & 6 & 8 \\
\hline
  \end{tabular}
  \caption{Conditional action probabilities for second two player \acha{} agent.}
\end{sidewaystable}